%% file: main.tex
\documentclass[authoryear,preprint,review,12pt]{elsarticle}
\usepackage{amssymb}
\input{preamable}
\journal{Neural Networks}
\begin{document}
\begin{frontmatter}
\title{Bayesian Disturbance Injection: \\Robust Imitation Learning of Flexible Policies for Robot Manipulation}

\author[inst1]{Hanbit Oh}

\affiliation[inst1]{organization={Division of Information Science, Graduate School of Science and Technology, NAIST},
            addressline={8916-5, Takayama-cho}, 
            city={Ikoma-city},
            postcode={630-0192}, 
            state={Nara},
            country={Japan}}

\author[inst1]{Hikaru Sasaki}
\author[inst1]{Brendan Michael}
\author[inst1]{Takamitsu Matsubara}

\begin{abstract}
Humans demonstrate a variety of interesting behavioral characteristics when performing tasks, such as selecting between seemingly equivalent optimal actions, performing recovery actions when deviating from the optimal trajectory, or moderating actions in response to sensed risks. However, imitation learning, which attempts to teach robots to perform these same tasks from observations of human demonstrations, often fails to capture such behavior. Specifically, commonly used learning algorithms embody inherent contradictions between the learning assumptions (\eg single optimal action) and actual human behavior (\eg multiple optimal actions), thereby limiting robot generalizability, applicability, and demonstration feasibility. To address this, this paper proposes designing imitation learning algorithms with a focus on utilizing human behavioral characteristics, thereby embodying principles for capturing and exploiting actual demonstrator behavioral characteristics. This paper presents the first imitation learning framework, Bayesian Disturbance Injection (BDI), that typifies human behavioral characteristics by incorporating model flexibility, robustification, and risk sensitivity. Bayesian inference is used to learn flexible non-parametric multi-action policies, while simultaneously robustifying policies by injecting risk-sensitive disturbances to induce human recovery action and ensuring demonstration feasibility. Our method is evaluated through risk-sensitive simulations and real-robot experiments (\eg table-sweep task, shaft-reach task and shaft-insertion task) using the UR5e 6-DOF robotic arm, to demonstrate the improved characterisation of behavior. Results show significant improvement in task performance, through improved flexibility, robustness as well as demonstration feasibility.
\end{abstract}


\end{frontmatter}

\section{Introduction}

\begin{figure}
    \centering
    \includegraphics[width=1.0\hsize]{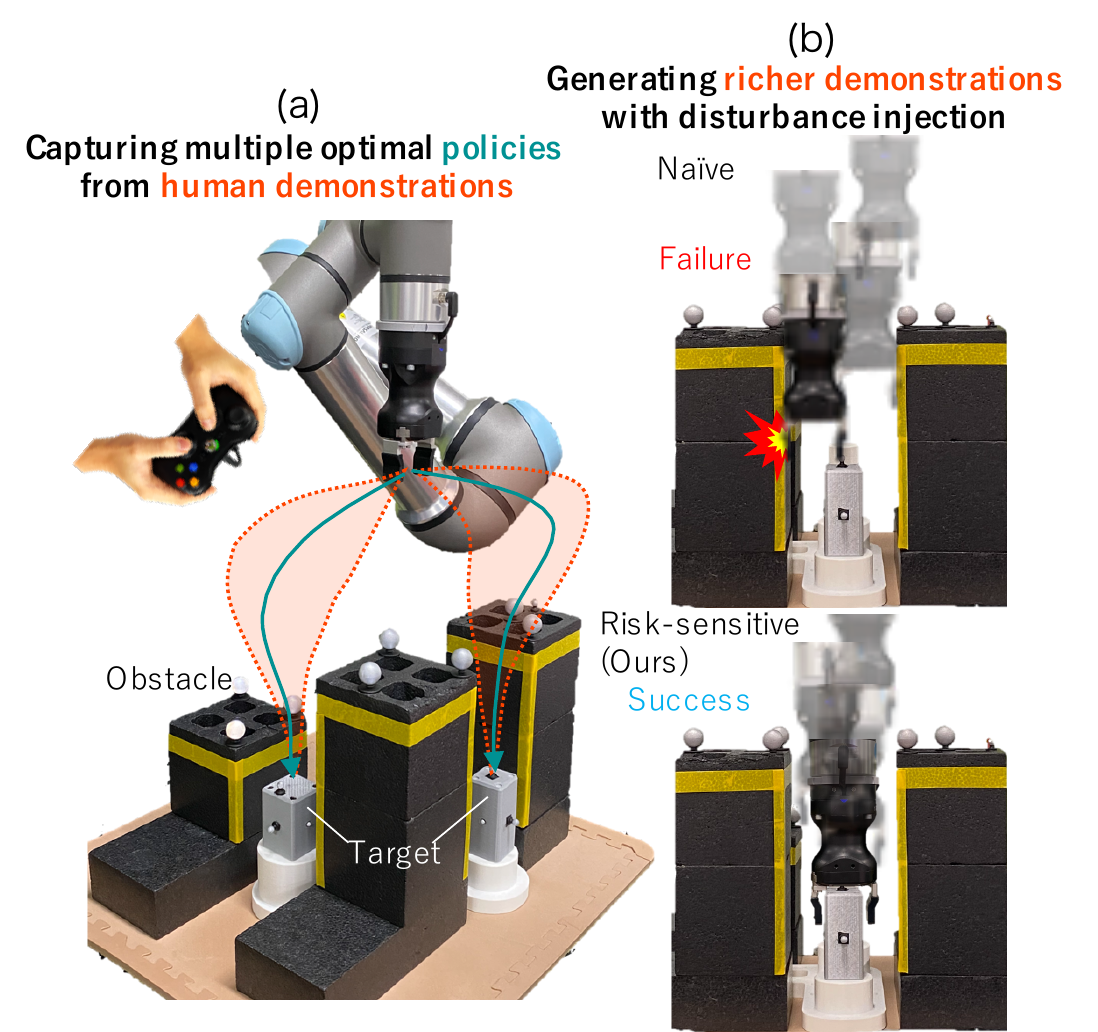}
    \caption{
{Illustration of the critical functions of the proposed method.}
(a): Multiple optimal policies are captured from complex human demonstrations, which may involve multiple optimal actions.
(b): Generating richer (more exploratory) demonstrations by injecting disturbances into expert's actions. 
Risk-sensitive disturbance models, which regulates its level response to risks of states, is employed to ensure demonstration feasibility.
}
    \label{fig:robot_manipulation}
\end{figure}
Creating robot controllers via machine learning has found widespread usage in both research \citep{kuindersma2016optimization, levine2018learning} and commercial \citep{levine2014learning,zhang2018deep, wang2021assembly,dadhich2016key,bojarski2016end} applications. Controller learning often utilizes large-scale exploration \citep{levine2018learning}, reward mechanisms (\eg an optimal control or reinforcement learning), and with highly accurate dynamics models \citep{kuindersma2016optimization} to learn autonomous control. However, in scenarios with exploration or dynamics model sparsity, imitation learning \citep{billard2008survey, argall2009survey, osa2018algorithmic} is an intuitive method for learning skills via observations of an expert demonstrator, avoiding complex explicit programming, reward design, or large-scale exploration.

Although many conventional imitation learning methods have been proposed, they are often inherently flawed when learning realistic robot manipulation tasks from humans. Specifically, methods often have fundamental algorithmic contradictions between the assumed characteristics of demonstration data, and the actual characteristics embodied by human demonstrators. For example, methods often algorithmically presuppose unique optimal actions for any given state \citep{pomerleau1989alvinn,levine2014learning,giusti2016forest}, or that sufficient exploratory actions can be performed in task space \citep{bojarski2016end,dyrstad2018fish}. However, in many practical scenarios, human demonstrators may act contrary to these assumptions, exhibiting behavior such as multiple optimal actions or performing actions in idiosyncratic patterns that lack exploratory movements.

To emphasise the significance of the problem, consider the task of learning robotic grasping \citep{cutkosky1990human, napier1956prehensile} with multiple targets or obstacles, whereby demonstrators provide demonstrations controlling a robot to grasp an object. Standard imitation learning algorithms presuppose traits such as unique optimal grasp configurations, and sufficient diversity of demonstrations of this unique configuration to ensure generalizability. However, human behavioral characteristics introduce uncertainty or probabilistic behavior, which challenges these assumptions. 
For example, demonstrators may arbitrarily or idiosyncratically select between various equivalently optimal actions to determine which path to take, or may be biased to specific regions of the demonstration space. This fundamental contradiction limits the generalizability of the policy by introducing additional modelling complexities, such as the covariate shift \citep{ross2010smile}.

Therefore, robotic manipulation in real-world scenarios necessitates the design of algorithms that embody principles for capturing actual \textit{demonstrator behavioral characteristics}. 
Specifically, this paper focuses on three key characteristics that are not typified by standard imitation learning algorithmic assumptions:
\begin{enumerate*}
    \item the ability to \textit{flexibly} adapt to a wide range of spatial scenarios,
    \item the capability to \textit{robustly} overcome deviation from optimal trajectories, and
    \item the ability to \textit{risk-sensitively} respond to ensure feasibility. 
\end{enumerate*}
As an illustrative example of these characteristics,  \fref{fig:robot_manipulation} demonstrates a robot reaching and grasping shafts while avoiding obstacles; where there are multiple seemingly equivalent optimal actions (green arrows). As such, this necessitates \textit{flexible policy models} capable of learning multiple optimal policies. Concurrently, robustification is applied using \textit{disturbance injection approaches} \citep{laskey2017dart,Oh2021} to expand demonstration coverage (shaded region) and induce the learning of recovery behavior for retaining the optimal trajectory. However, as shown in \fref{fig:robot_manipulation}-(b), na\"ive disturbance injection may result in demonstration infeasibility (\eg environmental collisions or confusion in decision making) thereby necessitating \textit{risk-sensitive disturbance models}, which regulate the disturbance level in response to state riskiness, thereby ensuring demonstration feasibility.

To the authors' knowledge, there is no unified framework for imitation learning that can simultaneously consider all of the above requirements. Our insight into this stems from the difficulty of formulating all three elements as a single framework. For example, in flexible policy learning, using non-parametric probabilistic policy models (\eg \citep{sasaki2019multimodal}) effectively captures multiple optimal actions from real human demonstrations where the number of optimal actions in each state cannot be specified a priori. However, the previously proposed disturbance injection method \citep{laskey2017dart} optimizes the disturbance level by minimizing the covariate shift, which corresponds to the maximum likelihood estimation based on the assumption of a deterministic policy model and a fixed disturbance level parameter. Thus, such flexible policy learning cannot be directly integrated into the previous framework. To address this difficulty, we propose to reformulate it as a non-parametric Bayesian inference problem, which employs the objective function of robustification as the likelihood and other non-parametric flexible policy and risk-sensitive disturbance models as the prior distribution. As such, this paper presents a novel Bayesian imitation learning framework that learns a probabilistic policy model capable of being both flexible to variations in demonstrations and robust to sources of error in policy application by injecting risk-sensitive disturbances, referred to as Bayesian Disturbance Injection (BDI).

\begin{figure}
    \centering
    \includegraphics[width=0.9\hsize]{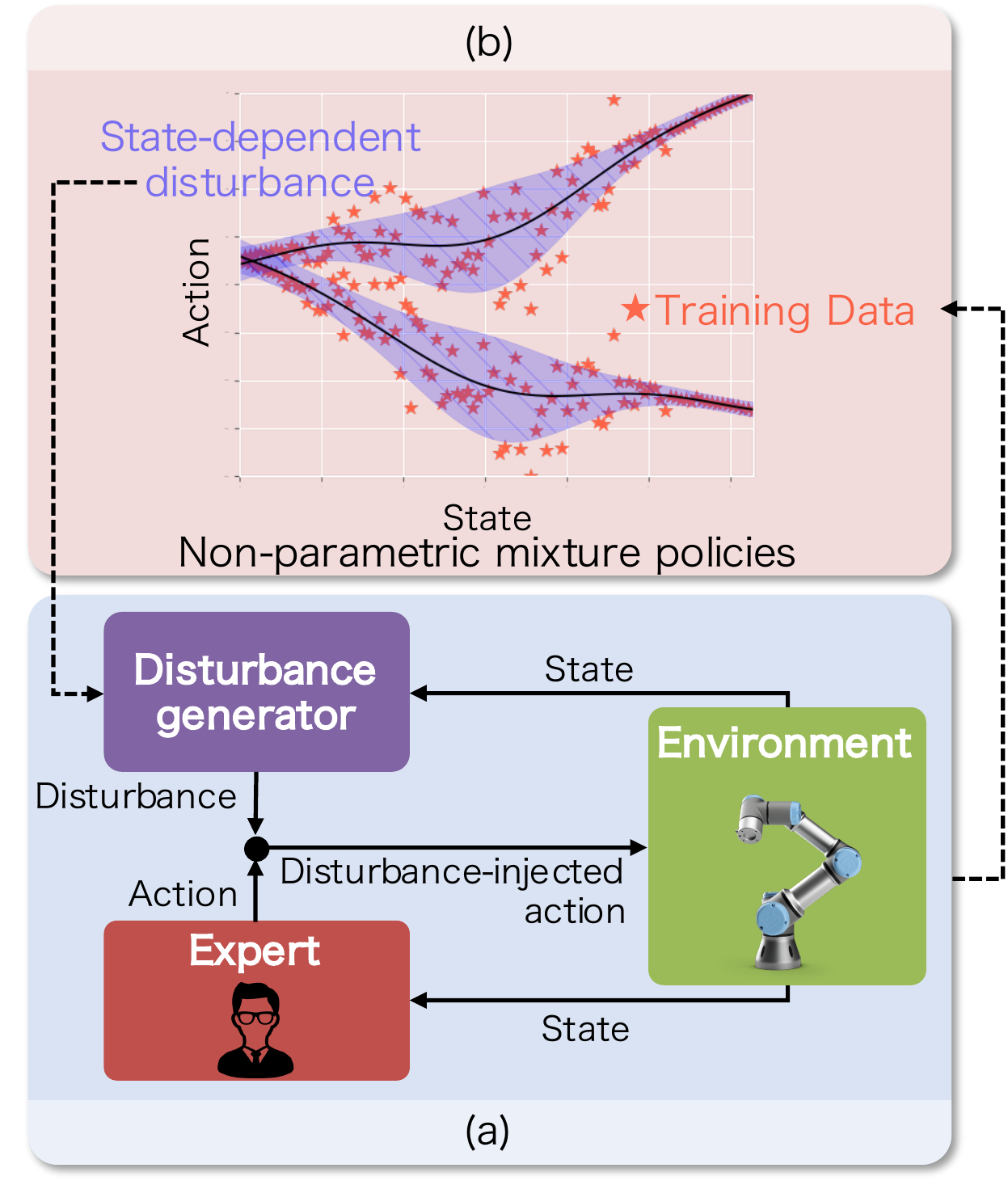}
    \caption{{Overview of MHGP-BDI, learning robust multi-modal policy with state-dependent disturbance injection.} (a): Collect and accumulate training datasets by injecting disturbance into the expert's demonstration actions. (b): Optimize the disturbance at which level can be regulated in a state-dependent manner. These processes (a) and (b) are repeated to obtain a robust multi-modal policy finally.}
    \label{fig:overview}
\end{figure}

Specifically, this paper establishes Multi-modal Heteroscedastic Gaussian Process BDI (MHGP-BDI), in which robust multi-modal probabilistic policy learning uses flexible regression models \citep{ross2013nonparametric} as non-parametric mixture policies (\fref{fig:overview}-(b)). To learn robustificition, the demonstrator is induced to provide recovery behavior, via disturbances injected into their actions (\fref{fig:overview}-(a)). To model risk-sensitive behavior, \textit{state-dependent disturbances} are learnt, which are approximated during policy learning via a heteroscedastic variance regression model \citep{lazaro2011variational} (\fref{fig:overview}-(b)). Given this conditional relationship between policy learning and disturbance optimization, this approach unifies learning within a single probabilistic framework. As such, inference of the policy and injection disturbance is performed simultaneously by variational Bayesian inference, thereby presenting a more authentic characterisation of the experts' behavior for imitation learning.

To evaluate the effectiveness of the proposed framework, experiments in learning probabilistic behavior from risk-sensitive simulation (\eg wall-avoidance task) and real robot experiments (\eg table-sweep task, shaft-reach task, shaft-insertion task) using the UR5e 6-DOF robotic arm are performed. Results show improved flexibility and robustness with increased learning performance and demonstration feasibility relative to comparison methods, giving a novel viewpoint of human behavioral characteristic learning.

Advancing from our preliminary publication \citep{Oh2021}, the key contributions of this paper are as follows:
\begin{enumerate}[\arabic*.]
    \item provides a novel perspective on imitation learning that captures integrated human behavior characteristics;
    \item provides a formulation that incorporates imitation learning models of flexibility, robustification, and risk sensitivity via a non-parametric Bayesian approach;
    \item provides a novel Bayesian imitation learning framework, Bayesian Disturbance Injection (BDI), to learn flexible non-parametric multi-action policies, while simultaneously robustifying policies by injecting risk-sensitive disturbances to induce human recovery action and ensuring demonstration feasibility;
    \item validates the effectiveness of the proposed approach by comparing it with state-of-the-art methods on simulations (\eg wall-avoidance task) and real robot experiments of general assembly tasks (\eg table-sweep task, shaft-reach task, shaft-insertion task);
\end{enumerate}

The remainder of this paper is organized as follows. \sref{sec:related_work} summarizes previous research on imitation learning. 
\sref{sec:preliminaries} introduces preliminaries of imitation learning from human demonstrations.
\sref{sec:Proposed_method} presents our proposed methods.
\sref{sec:Simulation} presents the simulation setup and results. 
\sref{sec:real_experiments} presents the experimental results in real robot experiments on assembly tasks. 
Finally, discussion and conclusion are described in Sections \ref{sec:discussion} and \ref{sec:conclusion}, respectively.

\section{Related Work}\label{sec:related_work}
A key objective of imitation learning is to ensure that models can capture the variation and stochasticity inherent in human motion while ensuring application of the learned policy can restrain deviation from optimal trajectories and ensure humans can accomplish demonstrations. To address these, prior approaches explore modelling \textit{flexibility}, and \textit{robustification} methods retain generalizability by mitigating compounding errors in policy application. However, na\"ive robustification may influence \textit{demonstration feasibility}, and as such, methods for addressing this trade-off are explored.

\subsection{Flexibility}\label{sec:relate:flex}
Learning generalized optimal action policies from human demonstrations, which often contain complex behaviors (\eg multiple optimal actions for a task), requires elaborate policy models with non-linearity and stochasticity.

Classical approaches to modelling uses dynamical frameworks for learning trajectories from demonstrations, \eg Dynamic Movement Primitives (DMPs), which can generalize the learned trajectories to new situations (\ie goal location or speed). However, this generalization depends on heuristics (\eg the appropriate number of basis functions regarding the complexity of trajectories), and is thus unsuitable for learning state-dependent feedback policies \citep{schaal2005dmp,ijspeert2013dynamical,khansari2011learning}.

To avoid imposing a priori structure, Gaussian Mixture Regression (GMR) is a non-parametric, intuitive means to learn trajectories or policies from demonstrators in the state-action-space. In this, Gaussian Mixture Modelling (GMM) \citep{calinon2016tutorial} is used as a basis function to capture non-linearities during learning, and has been utilized in imitation learning that deals with human demonstrations \citep{kyrarini2019robot}. However, GMR requires that basis functions be engineered by hand to deal with high-dimensional systems \citep{huang2019kernelized}.

In data-driven manner, nonlinearities can also be captured flexibly using Variational Auto-Encoders (VAE), which is a generative model that can embed high-dimensional features in latent variables \citep{kingma2013vae}. Furthermore, Conditional VAE (CVAE) can learn multi-modality by conditioning latent variables on a decoder \citep{sohn2015cvae}, and has been applied to capture multiple optimal actions from human demonstrations \citep{rahmatizadeh2018vision, ren2020generalization}.
However, such CVAE-based methods typically require large amounts of data to capture multi-modality, and even though such multi-modality is obtained using latent variables, learned policies may be sub-optimal for the high precision task; since latent variables are randomly sampled from a standard Gaussian prior distribution \citep{hsiao2019learning}.

As an alternative, Gaussian Process Regression (GPR) deals with implicit (high-dimensional) feature spaces with kernel functions. It thus can directly deal with high-dimensional observations without explicitly learning in this high-dimensional space \citep{rasmussen2003gaussian}. In particular, Overlapping Mixtures of Gaussian Processes (OMGP) \citep{lazaro2012overlapping} learns a multi-modal distribution by overlapping multiple GPs, and has been employed as a policy model with multiple optimal actions on flexible task learning of robotic policies \citep{sasaki2019multimodal}. To further reduce a priori tuning, Infinite Overlapping Mixtures of Gaussian Processes (IOMGP)  \citep{ross2013nonparametric} requires only an upper bound of the number of GPs to be estimated. As such, IOMGP is an intuitive means of learning flexible multi-modal policies from unlabeled human demonstration data and is employed in this paper.

\subsection{Robustness and Demonstration Feasibility}\label{sec:relate:robust}
While flexibility is key to capturing demonstrator motion, a major issue limiting application of learned policies is the problem of \textit{covariate shift} \citep{ross2010smile}. Specifically, environment variations (\eg manipulator starting position) induces differences between the policy distribution as learned by the manipulator and the actual task distribution during application.  

A more general approach to minimizing the covariate shift in imitation learning is Dataset Aggregation (DAgger) \citep{ross2011dagger}, whereby when the robot moves to a state not included in the training data, the expert augments the model by teaching the optimal recovery. 
However, this approach has limited applicability in practice due to the risk of exploring unknown states during policy application and the high overhead cost of human experts continuing to teach the robot the optimal actions.

An intuitive approach to robustifying learned policies against sources of error, without needing to a priori specify task-relevant learning parameters, is to exploit phenomenon similar to persistence excitation \citep{sastry2011adaptive}. In this, disturbances are injected into the expert's demonstrated actions, and the recovery behavior of the expert is learned given this perturbation. In an imitation learning context, Disturbances for Augmenting Robot Trajectories (DART) \citep{laskey2017dart} exploits this phenomenon for learning a deterministic policy model with a single optimal action. Additionally, DART is well suited to creating a richer dataset, by concurrently determining the optimal disturbance level to be injected into the demonstrated actions during policy learning.

However, the applicability of algorithms proposed to implement DART \citep{laskey2017dart} is limited, since DART employs a na{\"i}ve disturbance model which cannot regulate the level of disturbance regarding given states (\ie state-independent disturbance). For robotics tasks with local precision involving small clearance, such as a \fref{fig:robot_manipulation}, applying a uniform level of disturbance regardless of the state may lead to the risk of unintended collisions. On the other hand, simply limiting a level of disturbance to a small level does not sufficiently reduce the covariate shift.

A natural approach to addressing this in an imitation learning context, is to explore how human demonstrators approach this problem. Demonstrators, when aware of environmental risks, decrease movement velocity to increase action accuracy \citep{nagengast2011risk}, based on a \textit{speed-accuracy trade-off} \citep{wickelgren1977speed}. Inspired by such risk-sensitive behavior, this paper proposes a state-dependent disturbance model, which regulates the disturbance level to be small at risky states (\eg close to obstacles). As such, our disturbance injection robustifies policies, while maintaining demonstration feasibility. Specifically, a Heteroscedastic Gaussian Process (HGP) \citep{lazaro2011variational}, which can accurately infer probabilistic regression models with input-dependent variance, and is employed as a state-dependent disturbance model in this paper.

\section{Preliminaries}\label{sec:preliminaries}

\subsection{Imitation Learning from Expert's Demonstration}\label{sec:preliminaries:bc}
The objective of imitation learning is to learn a control policy by imitating the action from the expert's demonstration data.
A dynamics model is denoted as Markovian with a state $\mathbf s_t \in \mathbb{R}^{Q}$, an action $a_t \in \mathbb{R}$ , an initial state probability $p(\mathbf{s}_0)$ and a state transition distribution $p(\mathbf s_{t+1}\mid \mathbf s_t, a_t)$.
For simplicity but without loss of generality, the following derivation involves on one-dimensional action.
In this, a policy $\pi(a_t\mid\mathbf s_t)$ decides an action from a state, while a trajectory $\traj=(\mathbf{s}_{0},{a}_{0},\mathbf{s}_{1},{a}_{1}\dots {a}_{T-1},\mathbf{s}_{T})$ is a sequence of state-action pairs of $T$ steps.
The trajectory distribution is defined as:
\begin{align}
    p(\traj\mid\pi)=p(\mathbf{s}_0)\prod_{t=0}^{T-1}\pi(a_t\mid\mathbf{s}_t)p(\mathbf{s}_{t+1}\mid\mathbf{s}_t,a_t).
\end{align}

A significant aspects of imitation learning is to reproduce the expert's behavior, thus the function to compute the expected similarity between two policies with regard to trajectories is defined as:
\begin{align}
    J(\pi,\pi^{\prime}\mid \traj) = -\sum_{t=0}^{T-1} \mathbb E_{\pi(a\mid \mathbf s_t),\pi^{\prime}(a'\mid \mathbf s_t)}\left[||a - a'||_{2}^{2}\right].
\end{align}
A learned policy $\pi^R$ is obtained by solving the following optimization problem using a trajectory collected by an expert's policy $\pi^*$:
\begin{align}\label{eq:BC:policy}
    \pi^{R}= \argmax_\pi\mathbb{E}_{p(\traj\mid\pi^*)}\left[J(\pi,\pi^{*}\mid \traj)\right].
\end{align}

As discussed in \sref{sec:relate:robust}, such imitation learning may suffer from the problem of covariate shift, where the agent applying the learned policy drifts away from the demonstrated states due to compounding errors. This drift issue is delineated as the distributive difference between the trajectory during training data collection and learned policy application:
\begin{equation}
    \left|\mathbb{E}_{ p(\traj\mid\pi^*)}[J(\pi^{R},\pi^{*}\mid\traj)] -\mathbb{E}_{ p(\traj\mid\pi^R)}[J(\pi^{R},\pi^{*}\mid\traj)]\right|.
\end{equation}

\begin{figure}
    \centering
    \includegraphics[width=1.0\hsize]{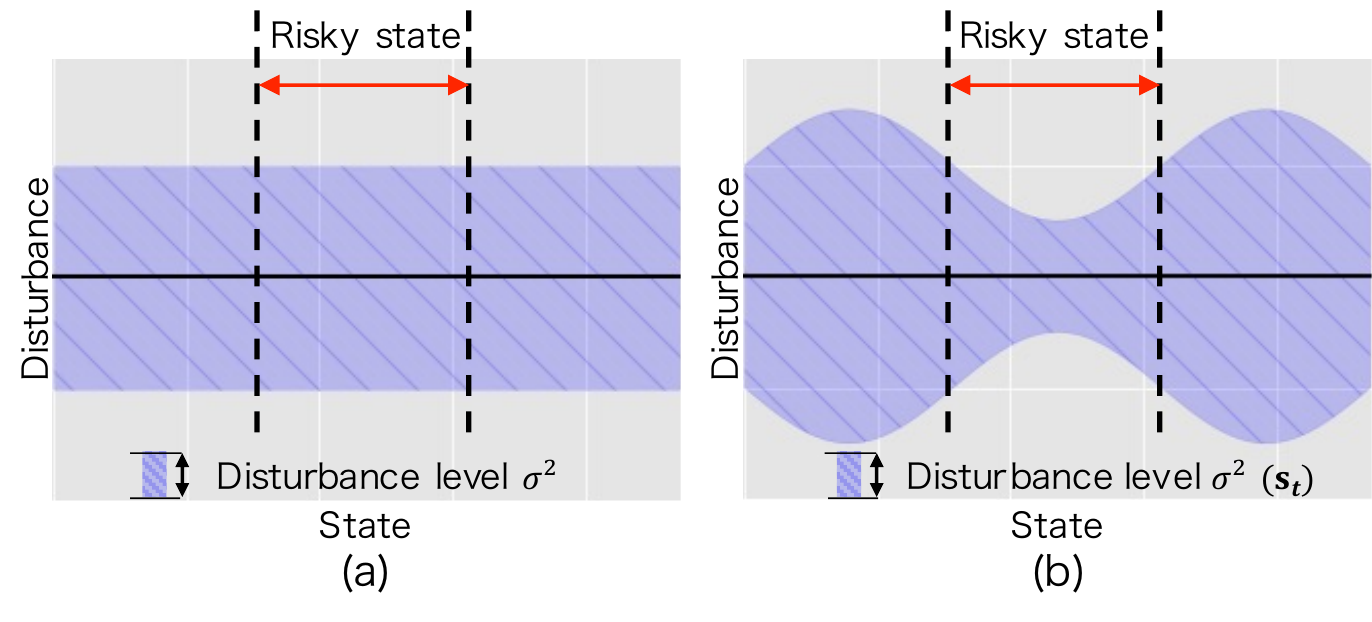}
    \vspace{-9mm}
    \caption{
    {Illustration of the comparison between (a) state-independent and (b) state-dependent disturbance models. }
    (a): a constant disturbance level regardless of the risk of the state, may be dangerous in risky states.
    (b): the disturbance level can be modified according to the state; \eg in risky states the disturbance level is reduced.
    }
    \label{fig:2type_disturbance}
\end{figure}

\subsection{Robust Imitation Learning by Injecting Disturbance into Expert}\label{sec:preliminaries:dart}
To learn robust policies from compounding errors, DART has been previously proposed \citep{laskey2017dart}. In this approach, disturbances are injected into the expert demonstrations to generate a richer training data set. The level of injecting disturbances is optimized, to reduce the covariate shift between the collected demonstration data and predicted trajectories.
The disturbance distribution is optimized iteratively in the data collection process.
Finally, the robust policy is learned using the collected data.

This injection disturbance is assumed that sampled from a Gaussian distribution as $\epsilon_t\sim\mathcal N(0, \sigma^2_k)$, where $k$ is the number of optimization iterations. The injection disturbance $\epsilon_t$ is added into the expert's action $a^*_t$.
The distribution of trajectories from a disturbance injected expert, is denoted as $p(\traj\mid\pi^*,\sigma^2_k)$ and the distribution of trajectories from a learned policy is $p(\traj\mid\pi^R_k)$.
To reduce the covariate shift, DART proposes to use the upper bound of covariate shift by Pinsker's inequality as:
\begin{align}\label{eq:DART:upper_covariate}
    &\left|\mathbb E_{p(\traj\mid\pi^*,\sigma^2_k)} \left[J(\pi^{R},\pi^{*}\mid\traj)\right]-\mathbb E_{p(\traj\mid\pi^R_k)} \left[J(\pi^{R},\pi^{*}\mid\traj)\right]\right|\nonumber\\
    &\leq T \sqrt{\frac{1}{2} \mathrm{KL}\left(p(\traj\mid\pi^R_k) \mid\mid p(\traj\mid\pi^*,\sigma^2_k)\right)},
\end{align}
where, $\mathrm{KL}(\cdot\mid\mid\cdot)$ is Kullback-Leibler divergence.
However, the upper bound \eref{eq:DART:upper_covariate} is analytically intractable to compute since the trajectory distribution of learned policy $p(\traj\mid\pi^R_k)$ is unknown.
Therefore, DART solves the upper bound by replacing the trajectory distribution of the learned policy with the trajectory distribution of the disturbance-injected expert. As such, data are collected over several iterations and a disturbance distribution is optimized at each iteration as:
\begin{align}\label{eq:DART:disturbance}
    &\sigma_{k+1}^{2} =\argmax_{\sigma^{2}} \mathbb{E}_{p(\traj\mid\pi^{*},\sigma^2_{k})} \nonumber \\
    &~~~~~\left[\sum_{t=0}^{T-1} \mathbb E_{\pi^R_{k}(a_t'\mid\mathbf s_t)} \left[\log \mathcal{N}\left( a_t'\relmiddle| a_t, \sigma^{2}\right)\right]\right], 
\end{align}
where, a learned policy at $k$~th iteration $\pi^R_k$ is obtained in the similar form as \eref{eq:BC:policy} by following:
\begin{align}\label{eq:DART:policy}
    \pi^R_k= \argmax_\pi\sum_{i=1}^{k-1}\mathbb{E}_{p(\traj\mid\pi^*,\sigma_i^2)}[J(\pi,\pi^{*}\mid\traj)].
\end{align}

Although DART can reduce the covariate shift by injecting disturbances into expert demonstrations, its applicability still suffers from the following issues.
The applied policy model is deterministic, which means that it cannot recognize complex human behavior (\eg multiple optimal actions) from the training data. 
Additionally, as shown in \fref{fig:2type_disturbance}-(a), the disturbance is injected uniformly regardless of the current state of the robot, which may induce dangerous situations (\eg physical contacts as in \fref{fig:robot_manipulation}-right). Furthermore, the disturbance level optimization \eref{eq:DART:disturbance} corresponds to the maximum likelihood estimation based on the assumption of a deterministic policy model and a fixed disturbance level parameter; thus, non-parametric policy learning (\eg \citep{sasaki2019multimodal}) in which effectively captures multiple optimal actions without requiring the specified number of optimal actions in each state, cannot be directly integrated into the DART framework. Therefore, a scheme to resolve these issues simultaneously via non-parametric Bayesian inference is derived in the next section.

\section{Proposed Method}\label{sec:Proposed_method}
In this section, a novel Bayesian imitation learning framework is proposed (\fref{fig:overview}) to learn a probabilistic policy via expert demonstrations with disturbance injection.
Specifically, flexibility, robustness, and risk-sensitivity are incorporated as a single formulation in a Bayesian manner; thus, it is referred to as Bayesian Disturbance Injection (BDI). The general form of BDI is derived in \sref{sec:Proposed_method:BDI}. As an overview, a non-parametric mixture model is utilized as a policy prior for capturing multiple optimal actions from human demonstration. A heteroscedastic model is employed as a disturbance prior for regulating disturbance level regarding states as shown in \fref{fig:2type_disturbance}-(b). The disturbance optimization term \eref{eq:DART:disturbance} is employed as a likelihood for minimising the covariate shift. This combination derives an imitation learning method, which learns a multi-modal policy and an injection disturbance distribution by Bayesian inference. Given this model, the predictive distribution is induced in a Bayesian form.
A specific implementation of BDI, which employ IOMGP \citep{ross2013nonparametric} as a policy prior and HGP \citep{lazaro2011variational} as a disturbance prior, is derived from \sref{sec:Proposed_method:MHGP-BDI}. 

\subsection{Bayesian Disturbance Injection (BDI)}\label{sec:Proposed_method:BDI}
Bayesian treatment is employed to learn probabilistic policies and disturbances in a single incorporated framework. As such, each goal function of the learning a policy \eref{eq:DART:policy} and disturbances \eref{eq:DART:disturbance} are formulated as a single likelihood. In addition, prior distributions of policy and disturbances are defined, and their respective posterior distributions are obtained via Bayesian inference.

To capture complex human behaviors as involving uncertainties, the probabilistic policy model which output action $a_t$ from the state $\mathbf{s}_t$ with Gaussian disturbance $\epsilon_t\sim\mathcal N(0, \sigma^2)$ is defined as: $a_t = f(\mathbf{s}_t) + \epsilon_t$, where $f(\cdot)$ is an output of a latent non-linear function. 
By applying this policy model to the objective function of policy learning \eref{eq:DART:policy}, a log-likelihood function that integrates policy and disturbances is derived as follows:
\begin{align}\label{eq:BDI:likelihood}
    J(\pi^{*},\mathbf{f}, \sigma^2 \mid \traj) 
    &= \sum_{t=0}^{T-1} \log p(a^*_t\mid f(\mathbf{s}_t),\sigma^2),
\end{align}
where, $\mathbf{f} = [f(\mathbf{s}_t)]_{t=0}^{T-1}$ is a set of a latent function outputs. Note that this log likelihood function \eref{eq:BDI:likelihood} is equal to the objective function of disturbance learning \eref{eq:DART:disturbance} if the mean and variables are swapped in a Gaussian distribution (the value of the distribution remains the same).

In addition, to infer a policy and disturbances in a non-parametric way from iteratively accumulated state-action pairs ($\{\mathbf{a}^*,\mathbf{S}\} = \{a^*_n, \mathbf{s}_n \}_{n=1}^{N}$, where $N = \sum_{j=1}^{k} N_j$ , $N_j$ is a size of the dataset that collected at $j$-th iteration), the prior distribution of a policy and disturbances are defined as $p(\mathbf{f}\mid\mathbf{S})$ and $p(\sigma^2)$, respectively.
Accordingly, posterior distributions of a policy and disturbances are simultaneously inferred by Bayesian inference as:
\begin{align}\label{eq:BDI:posterior}
    p(\mathbf{f} ,\sigma^2 \mid \mathbf{a}^*, \mathbf{S})= \frac{p(\mathbf{a}^* \mid \mathbf{f},\sigma^2)p(\mathbf{f}\mid \mathbf{S})p(\sigma^2)}{\pi^*(\mathbf{a}^*\mid\mathbf{S})}.
\end{align}
A summary of the BDI is shown in \aref{algorithm:BDI}.

\begin{algorithm}[tb]
\caption{BDI}
\label{algorithm:BDI}
\begin{algorithmic}[1]
\small
\renewcommand{\algorithmicrequire}{\textbf{Input:}}
\renewcommand{\algorithmicensure}{\textbf{Output:}}
\REQUIRE $\sigma_1^2$
\ENSURE {$p(\mathbf{f}, \sigma^2\mid \mathbf{a}^*, \mathbf{S})$}
\FOR {$k = 1$ to $K$}
    \STATE Get dataset through the disturbance injected expert:\\ $\{a^*_t,\mathbf{s}_t\}_{t=1}^{N_k} \sim p(\traj\mid\pi^*,\sigma_k^2)$
    \STATE Aggregate datasets :\\$\mathbf{a}^* \gets \mathbf{a}^* \cup \{a^*_t\}_{t=1}^{N_k}$ , $\mathbf{S} \gets \mathbf{S} \cup \{\mathbf{s}_t\}_{t=1}^{N_k}$ 
    \STATE Update $p(\mathbf{f}, \sigma^2\mid \mathbf{a}^*, \mathbf{S})$
\ENDFOR
\end{algorithmic} 
\end{algorithm}

\subsection{Multi-modal Heteroscedastic Gaussian Process BDI (MHGP-BDI)}\label{sec:Proposed_method:MHGP-BDI}
\subsubsection{Formulation:}
\begin{figure}
    \centering
    \includegraphics[width=1.0\hsize]{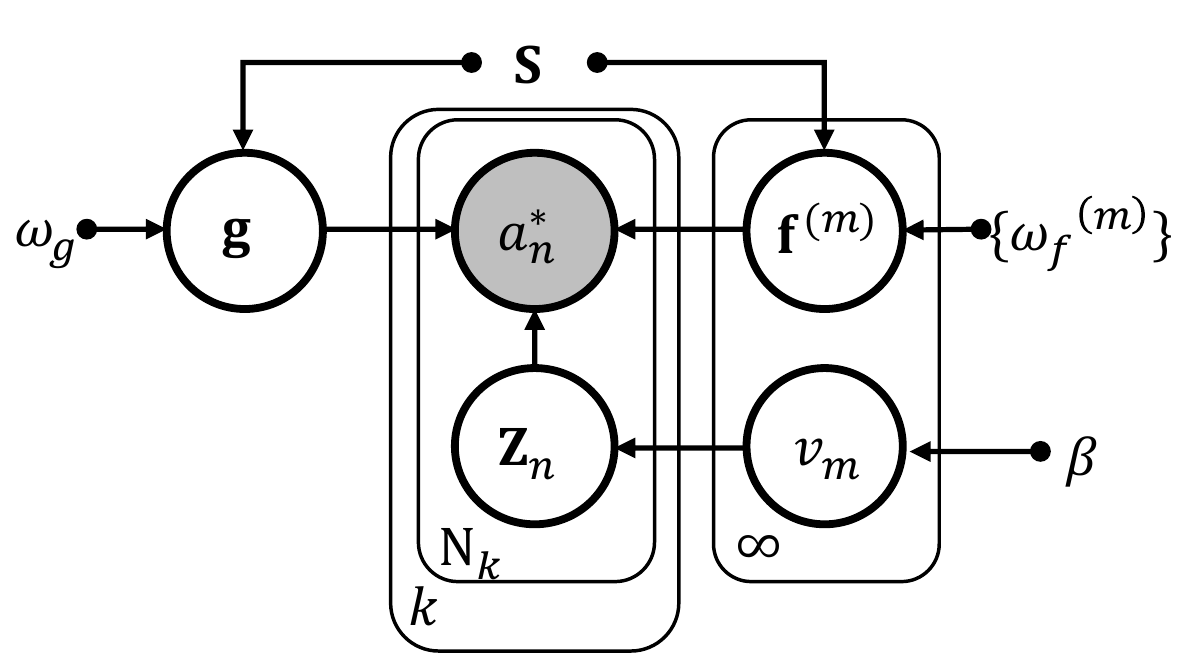}
    \caption{{Graphical model} of policy with state-dependent injection disturbance.
    }
    \label{fig:MHGP:grapicalmodel}
\end{figure}

To learn a multi-modal policy, the policy prior is considered as the product of infinite GPs, inspired by IOMGP. 
In addition, to learn state-dependent disturbances that can regulate its level respond to states, the prior of disturbances is considered as a state-dependent variance GP prior, inspired by HGP.
Intuitively, \fref{fig:MHGP:grapicalmodel} shows a probabilistic policy model in which expert's actions $\mathbf{a^*}$ are estimated by $\mathbf{f}^{(m)}, \mathbf{Z}, \mathbf{g}$.
The latent function $\mathbf{f}^{(m)}$ is the output of $m$-th GP given state $\mathbf{S}$. 
To allocate the expert's $n$-th action $a_n^*$ to the $m$-th latent function $\mathbf{f}^{(m)}$, the indicator matrix $\mathbf{Z} \in \mathbb{R}^{N\times \infty}$ is defined.
To estimate the optimal number of GPs, a random variable $v_m$ quantifies the uncertainty assigned to $\mathbf{f}^{(m)}$. 
In addition, to learn an injection disturbance which can regulate its level in a state-dependent way, a state-dependent disturbance level $\sigma^2(\mathbf{s}_n) = e^{g(\mathbf{s}_n)}$ is introduced, where $g(\cdot)$ is an output of GP given a state $\mathbf{s}_n$.

\textbf{Policy prior:}
the set of latent functions is denoted as $\{ {\mathbf{f}}^{(m)}\}=\{\mathbf{f}^{(m)}\}^{\infty}_{m=1}$ and a GP prior is given by :
\begin{equation}\label{eq:BDI:prior_f}
    p(\{\mathbf{f}^{(m)}\} \mid\mathbf{S}, \{\boldsymbol\omega^{(m)}\})=\prod_{m=1}^{\infty}  \mathcal{N}(\mathbf{f}^{(m)} \mid\mathbf{0},\Kbf^{(m)}; \omega^{(m)}),
\end{equation}
where $\Kbf^{(m)}=\mathrm{k}_{\mathbf{f}}^{(m)}({\mathbf{S}}, {\mathbf{S}})$ is the $m$-th kernel Gram matrix with the kernel function $\mathrm{k}_{\mathbf{f}}^{(m)}(\cdot,\cdot)$ and a kernel hyperparameter $\omega_{\mathbf{f}}^{(m)}$.
Let $\{\boldsymbol{\omega}_{\mathbf{f}}^{(m)}\} = \{\omega_{\mathbf{f}}^{(m)}\}^{\infty}_{m=1}$ be the set of hyperparameters of infinite number of kernel functions.

To infer the optimal number of GPs from the above GP mixtures \eref{eq:BDI:prior_f}, the Stick Breaking Process (SBP) \citep{sethuraman1994constructive} is used as a prior of $\mathbf{Z}$, which can be interpreted as an infinite mixture model as follows:
\begin{align}\label{prior_z}
    p(\mathbf{Z} \mid \mathbf{v}) &= \prod_{n=1}^{N} \prod_{m=1}^{\infty}\left(v_{m} \prod_{j=1}^{m-1}\left(1-v_{j}\right)\right)^{\mathbf{Z}_{nm}}, \\
    p(\mathbf{v} \mid \beta)&=\prod_{m=1}^{\infty} \operatorname{Beta}\left(v_{m} \relmiddle| 1,\beta\right).
\end{align}
Note that the implementation of variational Bayesian learning approximates infinite-dimensional inference with a predefined upper bound of $M$.
In this process, ${{v}_m}$ is a random variable indicating the probability that the data corresponds to the $m$-th GP. Thus, it is possible to estimate the optimal number of GPs with a high probability of allocation starting from an infinite number of GPs. $\beta$ is a hyperparameter of SBP denoting the level of concentration of the data in the cluster.

\textbf{Disturbance prior:}
the above policy model differs from the IOMGP model for regression \citep{ross2013nonparametric}; our model employs a state-dependent disturbance level $e^{g(\mathbf{s}_n)}$ where the values are determined in response to the state.
To learn a state-dependent disturbance, the disturbance prior is considered as a heteroscedastic Gaussian disturbance, inspired by HGP \citep{lazaro2011variational}. Accordingly, a GP prior is placed on a latent function $\mathbf{g} = \{g(\mathbf{s}_n)\}_{n=1}^{N}$, which represent a level of disturbance as:
\begin{align}\label{prior_g}
p(\mathbf{g} \mid\mathbf{S}; \omega_{\mathbf{g}})
=\mathcal{N}(\mathbf{g}\mid\mu_0\boldsymbol{1}_{N},\Kbg; \omega_{\mathbf{g}}),
\end{align}
where, $\mu_{0}$ is mean of disturbance distribution, $\mathbf 1_{N_i}$ is a vector whose size is $N_i$ and all components are one, and $\Kbg$ is kernel Gram matrix with a kernel hyperparameter $\omega_{\mathbf{g}}$.

\textbf{Likelihood:}
the likelihood function, as in \eref{eq:BDI:likelihood}, for the variables ($\{\mathbf{f}^{(m)}\}, \mathbf{g}, \mathbf{Z}$) in the policy and disturbance models, is derived as follows:
\begin{align}\label{MHGP:likelihood}
    &p({\mathbf{a}^*} \mid \mathbf{g}, \{\mathbf{f}^{(m)}\}, \mathbf{Z})\nonumber\\
    &=\prod_{n=1}^{N}\prod_{m=1}^{\infty} \mathcal{N}({a}_n^*\mid \mathbf{f}_n^{(m)},e^{\mathbf{g}_{n}})^{\mathbf{Z}_{nm}}.
\end{align}
This formulation is described in a graphical model that defines the relationship between the variables as shown in \fref{fig:MHGP:grapicalmodel}, and the joint distribution of the model as :
\begin{align}\label{MHGP:joint_distribution}
    &p({\mathbf{a}^*}, \mathbf{g}, \{\mathbf{f}^{(m)}\}, \mathbf{Z}, \mathbf{v}\mid\mathbf{S} ;\Omega ) \nonumber\\
    &=p({\mathbf{a}^*} \mid \mathbf{g}, \{\mathbf{f}^{(m)}\}, \mathbf{Z}) p(\mathbf{g} \mid\mathbf{S}; \omega_{\mathbf{g}})\cdot \nonumber\\
    &~~~~~~p(\{\mathbf{f}^{(m)}\} \mid\mathbf{S}; \{\boldsymbol{\omega}_{\mathbf{f}}^{(m)}\})
    p(\mathbf{Z} \mid \mathbf{v}) p(\mathbf{v} \mid \beta),
\end{align}
where $\Omega =(\{\boldsymbol{\omega}_{\mathbf{f}}^{(m)}\},{\omega}_{\mathbf{g}},\mu_0,\beta)$ represents a set of hyperparameters.

\subsubsection{Optimization of Policies and Injection Disturbance via Variational Bayesian Inference:}
Bayesian inference is a framework that estimates the posterior distributions of the policies and their predictive distributions for new input data rather than point estimates of the policy parameters.
To obtain the posterior and the predictive distributions, the marginal likelihood is calculated as :
\begin{align}\label{MGP:marginal_likelihood}
    &p({\mathbf{a}^*}\mid\mathbf{S};\Omega )\nonumber\\
    &= \int p({\mathbf{a}^*}, \mathbf{g}, \{\mathbf{f}^{(m)}\}, \mathbf{Z}, \mathbf{v}\mid\mathbf{S} ;\Omega )
    \dd\mathbf{g} \dd\{\mathbf{f}^{(m)}\} \dd \mathbf{Z} \dd \mathbf{v}.
\end{align}

However, it is intractable to calculate the log marginal likelihood of \eref{MGP:marginal_likelihood} analytically.
Therefore, the variational lower bound is derived as the objective function of variational learning. 
The true posterior distribution is approximated by the variational posterior distribution, which maximizes the variational lower bound.
Such the variational lower bound $\mathcal{L}(q,\Omega)$ is derived by applying the Jensen inequality to the log marginal likelihood, as:
\begin{align}\label{MHGP:variational_lower_bound}
    &\log {
    p({\mathbf{a}^*}\mid\mathbf{S};\Omega)
    }\nonumber\\
   &\geq  \int q\log 
   \frac{p({\mathbf{a}^*}, \mathbf{g}, \{\mathbf{f}^{(m)}\}, \mathbf{Z}, \mathbf{v} \mid\mathbf{S};\Omega)}{q} \dd \mathbf{g} \dd\{\mathbf{f}^{(m)}\} \dd \mathbf{Z} \dd \mathbf{v}\nonumber\\
   &=\mathcal{L}(q, \Omega),
\end{align}
where, $q=q(\mathbf{g, \{{f}^{(m)}\}, Z, v})$ represents a set of variational posteriors.

As a common fashion of variational inference, the variational posterior distribution is assumed to be factorized among all latent variables (known as the \textit{mean-field approximation} \citep{parisi1988statistical}) as follows:
\begin{equation}
    q(\mathbf{g, \{{f}^{(m)}\}, Z, v}) = q(\mathbf {g})q(\mathbf {f}^{(m)}) q(\mathbf{Z}) \prod_{m=1}^\infty q(v_m).
\end{equation}
In addition, to compute the variational lower bound in closed form, the posterior of $\mathbf{g}$ is restricted to a multivariate Gaussian distribution. Furthermore, to reduce the computational complexity and facilitate the optimization problem, similar to Gaussian approximation \citep{opper2009variational}, a positive variational parameter $\mathbf{\Lambda} = \mathrm{diag}\{\lambda_n\}_{n=1}^{N}$ is employed as :
\begin{align}
    q(\mathbf{g}) &= \mathcal{N}(\mathbf{g}\mid \boldsymbol{\mu}_{\mathbf{g}},\boldsymbol{\Sigma}_{\mathbf{g}}),\\
    {\boldsymbol{\mu}}_{\mathbf{g}}&=\Kbg\left(\mathbf{\Lambda}-\frac{1}{2} \mathbf{I}\right) \mathbf{1}_{N}+\mu_{0} \mathbf{1}_{N}, \\
    \quad {\boldsymbol{\Sigma}}_{\mathbf{g}}^{-1}&=\Kbg^{-1}+\mathbf{\Lambda},
\end{align}
where, $\mathbf{I}$ is an identity matrix.

Therefore, the optimization formulation is derived using the \textit{Expectation-Maximization} (EM)-like algorithm.
The variational posterior distributions $q$ are optimized with fixed hyperparameters $\Omega^{\prime}$ in E-step, and the hyperparameters $\Omega^{\prime}$ are optimized with fixed variational posterior distributions $q$ in M-step with:
\begin{align}\label{eq:MHGP:objective_function}
    \hat{q},\hat{\Omega}^{\prime} = \argmax_{q,\Omega^{\prime}}
    \mathcal{L}(q,\Omega^{\prime}),
\end{align}
where, $\Omega^{\prime}= (\Omega, \mathbf{\Lambda})$ represents a set of variational hyperparameters.
See \ref{appendix:update_q} for details of $q$ update laws and \ref{appendix:lowerbound} for details of lower bound of marginal likelihood.
In addition, a summary of the proposed method is shown in \aref{algorithm:MHGP-BDI}; 
and \tref{table:computation} shows the computational complexity of each optimization.

\begin{algorithm}[tb]
\caption{MHGP-BDI}
\label{algorithm:MHGP-BDI}
\begin{algorithmic}[1]
\small
\renewcommand{\algorithmicrequire}{\textbf{Input:}}
\renewcommand{\algorithmicensure}{\textbf{Output:}}
\REQUIRE $M, \sigma^2_1$
\ENSURE {$\hat{q},\hat{\Omega}^{\prime}$}
\FOR {$k = 1$ to $K$}
    \STATE Get dataset through the disturbance injected expert:\\ $\{a^*_t,\mathbf{s}_t\}_{t=1}^{N_k} \sim p(\traj\mid\pi^*,\sigma^2_{k})$
    \STATE Aggregate datasets :$\mathcal{D} \gets \mathcal{D}\cup \{a^*_t,\mathbf{s}_t\}_{t=1}^{N_k}$ 
\WHILE {$ \mathcal{L}(q,\Omega^{\prime})$ is not converged}
    \WHILE{$ \mathcal{L}(q,\Omega^{\prime})$ is not converged}
    \STATE Update $q(\mathbf{f}^{(m)})$, $q(\mathbf{Z})$,and $q(v_m)$ alternately
    \ENDWHILE
    \STATE{Optimize $ \Omega^{\prime}$ with fixed $q$:\\} $\hat{\Omega}^{\prime}\gets \argmax_{\Omega^{\prime}} \mathcal{L}(q,\Omega^{\prime})$
\ENDWHILE
\ENDFOR
\end{algorithmic} 
\end{algorithm}

\begin{table}[tb]
  \centering
  \caption{Computational complexity of each optimization in MHGP-BDI: $N$ and $M$ are number of training data sets and upper bound of mixtures, respectively. }
  \label{table:computation}
  \vspace{2mm}
  \begin{tabular}{|c|ccc|}\hline
    &$q(\mathbf{g})$ & $q(\mathbf{v})$ & $q(\{\mathbf{f}^{(m)}\})$, $q(\mathbf{Z})$, $\mathcal{L}$   \\ \hline\hline
    \multicolumn{1}{|c|}{\begin{tabular}[c]{@{}c@{}}MHGP-BDI\end{tabular}}
    &$\mathcal{O}(N^3)$ & $\mathcal{O}(M^2N)$ & $\mathcal{O}(MN^3)$ \\ \hline
  \end{tabular}
\end{table}

\subsubsection{Predictive Distribution:} Using variational parameter $\mathbf{\Lambda}$ optimized by maximizing \eref{eq:MHGP:objective_function}, the predictive disturbance $q(g_{*})$ on a new state $\mathbf{s}_{*}$ can be obtained as:
\begin{align}
q(g_{*})
&=\int p(g_{*} \mid \mathbf{s}_{*}, \mathbf{S}, \mathbf{g}) q(\mathbf{g}) \mathrm{d} \mathbf{g}\nonumber\\
&=\mathcal{N}(g_{*} \mid \mu_{g*}, \sigma_{g*}^{2}),\\
\mu_{{g_*}}
&=\mathbf{k}_{g *}^{\top}(\mathbf{\Lambda}- \mathbf{I}/2) \mathbf{1}_{N}+\mu_{0},\\
\sigma_{g_*}^2
&=k_{g * *}-\mathbf{k}_{g *}^{\top}(\Kbg+\mathbf{\Lambda}^{-1})^{-1} \mathbf{k}_{g *},
\end{align}
where $\mathbf{k}_{g *} = \mathrm{k}_{g}(\mathbf{s}^{*}, \mathbf{S})$, and $\mathrm{k}_{g * *} = \mathrm{k}_{g}(\mathbf{s}^{*}, \mathbf{s}^{*})$.
As such, a level of disturbance injected at the next iteration $k+1$ is calculated as: $\sigma_{k+1}^2(\mathbf{s}_{*}) = e^{\mu_{g_{*}}}$.

\begin{table}[tb]
  \centering
  \caption{Computational complexity of each prediction in MHGP-BDI: $N$ is number of training data sets.}
  \label{table:computation:pred}
  \vspace{2mm}
  \begin{tabular}{|c|cc|}\hline
    &$q(g_{*})$ & $p(a_{*}^{(m)} \mid \mathbf{s}_{*}, \mathbf{S}, \mathbf{a}^*)$  \\ \hline\hline
    \multicolumn{1}{|c|}{\begin{tabular}[c]{@{}c@{}}MHGP-BDI\end{tabular}}
    &$\mathcal{O}(N^3)$ & $\mathcal{O}(N^3)$ \\ \hline
  \end{tabular}
\end{table}

In addition, using the hyperparameters $\Omega^{\prime}$ and the variational posterior distributions $q$ optimized by variational Bayesian learning, the predictive distribution of the $m$-th action $a_{*}^{(m)}$ on a current state $\mathbf{s}_{*}$ is derived as:
\begin{align}\label{eq:predict:action}
&p(a_{*}^{(m)} \mid \mathbf{s}_{*}, \mathbf{S}, \mathbf{a}^*)\nonumber\\
& \approx \int p(a^*_{*} \mid \mathbf{f}^{(m)}, {g}_{*}, \mathbf{s}_{*}) q(\mathbf{f}^{(m)}) q({g}_{*}) \mathrm{d} \mathbf{f}^{(m)} \mathrm{d} {g}_{*}\nonumber\\
&= \int \mathcal{N}(a^*_{*} \mid \mu_{*}^{(m)}, c_{*}^{2(m)}+\exp({g}_*))
\mathcal{N}({g}_{*} \mid {{\mu}}_{{g}*}, {\sigma}^2_{{g}_*})
\mathrm{d} {g}_{*};
\end{align}
however, it is analytically intractable to compute. Alternatively, using a Gauss-Hermite quadrature rule \citep{liu1994Gaussquad}, mean $\mu_{*}^{(m)}$ and variance $\sigma_{*}^{2(m)}$ of the predictive distribution \eref{eq:predict:action} can be approximated as:
\begin{align}
    \mu_{*}^{(m)}
    &=\mathbf{k}^{(m)\top}_{f_*} (\Kbf^{(m)} + \mathbf{R}^{-1})^{-1}\mathbf{a}^*,\\
    \sigma_{*}^{2(m)}
    &= c_{*}^{2(m)}+\exp(\mu_{g_*}+{\sigma_{g_*}^2/2}),\\
    c_{*}^{2(m)}
    &=k_{f * *}^{(m)}-\mathbf{k}_{f *}^{(m)\top}(\Kbf^{(m)}+\mathbf{R}^{-1})^{-1} \mathbf{k}_{f *}^{(m)},
\end{align}
where $\mathbf{k}_{f *}^{(m)} = \mathrm{k}_{f}^{(m)}(\mathbf{s}^{*}, \mathbf{S})$, and $\mathrm{k}_{f * *}^{(m)} = \mathrm{k}_{f}^{(m)}(\mathbf{s}^{*}, \mathbf{s}^{*})$; and \tref{table:computation:pred} shows the computational complexity of each prediction.
Additionally, $m$ is chosen as the value that maximizes the inverse of the predicted variance $\sigma_{*}^{2(m)}$ as:
\begin{equation}
    \hat{m} = \underset{m}{\argmax}\frac{1}{\sigma_{*}^{2(m)}},
\end{equation}
as such, meaning the $\hat{m}$-th GP is selected, due to its minimal uncertainty.

\section{Simulation}\label{sec:Simulation}

In this section, the proposed methodology (MHGP-BDI) is evaluated in regards to the following questions, to examine key objectives of capturing human behavior characteristics in a simulated precision wall-avoidance task: 
\begin{enumerate*}
    \item flexibility: how does capturing multiple optimal human actions affect imitation learning of robotic tasks?, 
    \item robustness: how does injecting disturbances into human demonstrations affect the applicability of learned policies?, and 
    \item risk-sensitivity: how does injecting disturbance into human action command affect human demonstrations' feasibility?
\end{enumerate*}

\textbf{Evaluation Metrics: }
Performance of MHGP-BDI is considered during the training phase and execution phase. For the former, 
\textit{demonstration feasibility}, or the success rate of collecting training data with a human expert in the loop, is evaluated. On the latter, 
\textit{execution performance}, or the success rate of deploying the learned policy after training, is evaluated. 
These metrics are reported in the wall-avoidance simulation study (\sref{sec:Simulation:wall}) and the real robot assembly study (\sref{sec:real_experiments}). By comparing both performances across different algorithms, each algorithm is evaluated for how effectively it obtains policy performance while ensuring the demonstration feasibility.

\begin{table}[ht]
\centering
\caption{Comparison models in terms of flexibility, robustness, and demonstration feasibility.}
\label{table:comparisons}
\small
\begin{tabular}{|p{5cm}|C{2.cm}|C{2.cm}|C{3cm}|}
\hline
\multirow{2}{*}{Learning Models}  & \multirow{2}{*}{Flexibility} & \multirow{2}{*}{Robustness} & \multirow{1}{*}{Demonstration}\\
   &             &            &Feasibility\\\hline\hline
BC \citep{bain1995framework}   & \xmark      &\xmark      &\cmark  \\ \hline
DART \citep{laskey2017dart}   & \xmark      &\cmark      &\xmark  \\ \hline
CVAE-BC \citep{ren2020generalization}   & \cmark      &\xmark      &\cmark  \\ \hline
UGP-BC   & \xmark      &\xmark      &\cmark  \\ \hline
UGP-BDI  & \xmark      &\cmark      &\xmark  \\ \hline
UHGP-BDI & \xmark      &\cmark      &\cmark  \\ \hline
MGP-BC   & \cmark      &\xmark      &\cmark  \\ \hline
MGP-BDI \citep{Oh2021}  & \cmark      &\cmark      &\xmark  \\ \hline
\rowcolor[HTML]{C0C0C0} 
MHGP-BDI (Proposed)&{\cmark} &{\cmark}&{\cmark}\\\hline
\end{tabular}
\end{table}

\textbf{Comparison Methods: }
To evaluate the proposed method (MHGP-BDI), comparisons are made between 8 baselines. Each baseline's features (flexibility, robustness, and demonstration feasibility) are represented in \tref{table:comparisons}. Specifically, these algorithms are implemented as:
\begin{itemize}
    \item \textbf{Behavior Cloning (BC)} \citep{bain1995framework}: Conventional supervised imitation learning as described in \sref{sec:preliminaries:bc} using a neural network policy model,
    \item \textbf{Disturbances for Augmenting Robot Trajectories (DART)} \citep{laskey2017dart}: Robust imitation learning by injecting disturbance into expert as described in \sref{sec:preliminaries:dart} using a neural network policy model,
    \item \textbf{Conditional Variational AutoEncoders BC (CVAE-BC)} \citep{ren2020generalization}: Multi-modal imitation learning based on BC algorithm using a CVAE policy model,
    \item \textbf{Uni-modal GP Behavior Cloning (UGP-BC)}: BC using standard uni-modal GPs \citep{rasmussen2003gaussian},
    \item \textbf{Multi-modal GP BC (MGP-BC)}: BC using infinite overlapping mixtures of GPs (IOMGP),
    \item \textbf{UGP-BDI}: BDI using  standard uni-modal GPs and state-independent disturbance model with a constant disturbance level of $\sigma^2$,
    \item \textbf{Uni-modal Heteroscedastic GP BDI (UHGP-BDI)}: BDI using standard uni-modal GPs and Heteroscedastic Gaussian Processes (HGP) as state-dependent disturbance model $\sigma^2(\mathbf{s}_t)$,
    \item \textbf{MGP-BDI} \citep{Oh2021}: BDI using IOMGP policy model and state-independent disturbance model which level parameter as $\sigma^2$.
\end{itemize}
See \ref{appendix:hyper} for how the hyperparameters of each method are set.
Note, in all experiments, demonstrations are performed without injecting disturbances in the first iteration (\ie $\sigma^2_1 = 0$); since initially, there is no available evidence of which level of disturbance is suitable.

\subsection{Wall-avoidance Task}\label{sec:Simulation:wall}
Initially, a wall-avoidance task involving multiple apertures is presented (\fref{fig:sim:wide_env&multi}-(a)).
In this experiment, demonstrations are conducted in an environment involving states in which physical contact (\eg collisions of an agent and walls) is likely to occur, and the demonstration feasibility (\eg avoiding collision) will be evaluated.
The learned policy is evaluated through test execution episodes to evaluate its flexibility capturing multiple optimal actions from demonstrations (\eg multiple paths through an aperture to reach the goal), and robustness against environmental variations (\eg starting positions of the agent or inertial of the agent) that may induces the covariate shift.

\subsubsection{Setup}\label{sec:sim:setup}
\begin{figure}
    \centering
    \includegraphics[width=1.0\hsize]{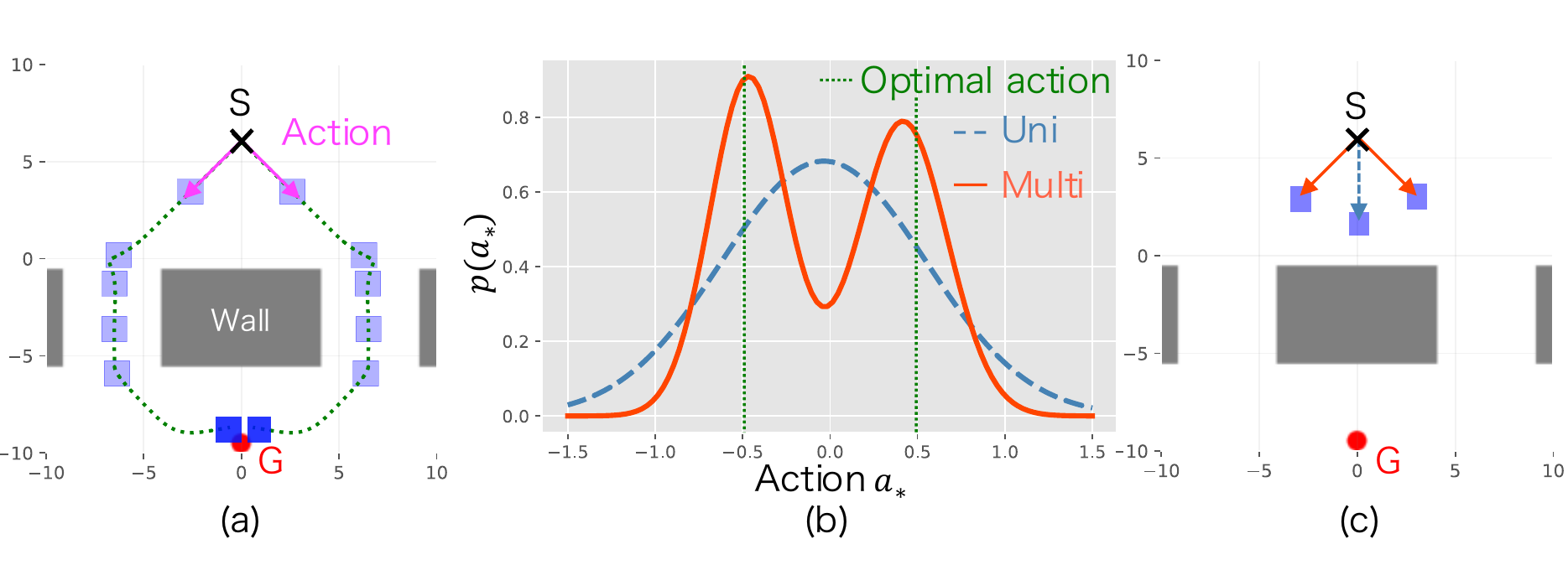}
    \vspace{-2mm}
    \caption{
    {Wall-avoidance Task (Wide).} (a): Environment of passing through a multiple aperture. S and G represent a starting and a goal position, respectively. An algorithmic supervisor's demonstrated movement, which includes the cautious phase (\eg move slow when a robot is close to an aperture), is captured as multiple frames with a $0.025$ frame rate. (b), (c): Comparing flexibility between multi-modal approaches and uni-modal approaches. (b) The predictive distribution of x-axis action $a_{*}$ in a given starting position state. (c) Movements of multi-modal approaches and uni-modal approaches at policy application phase.}
    \label{fig:sim:wide_env&multi}
\end{figure}

\begin{figure}
    \centering
    \includegraphics[width=\hsize]{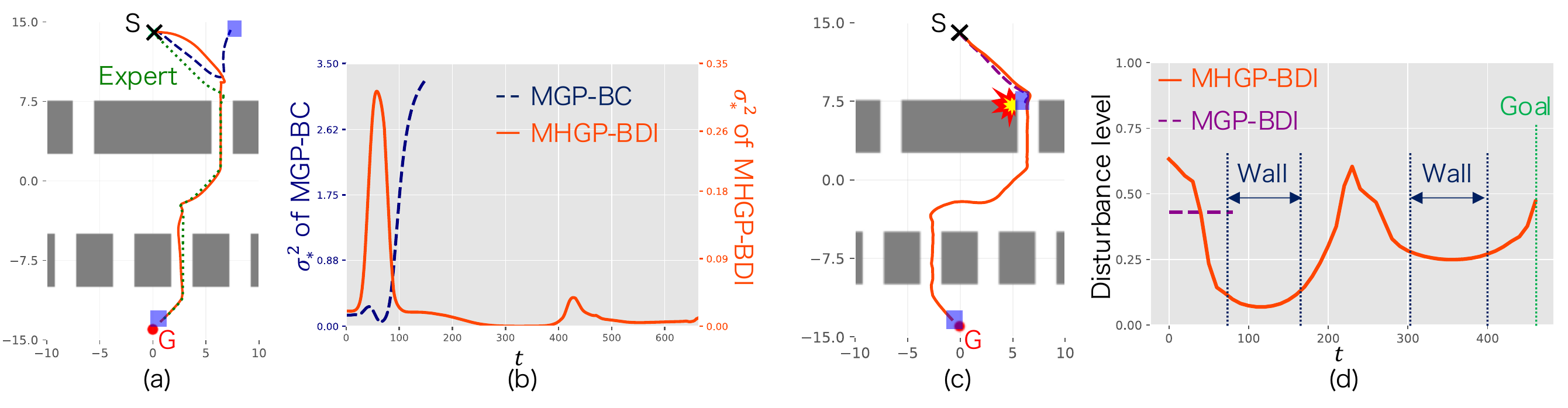}
    \caption{
    {Wall-avoidance Task (Complex).} (a), (b): Comparing robustness between MHGP-BDI and MGP-BC. (a) Generated  trajectory from policy learned by MHGP-BDI and MGP-BC, and (b) sequentially depicts the predictive action variance $\sigma_{*}^2$ (\ie norm of the XY-axis $\sigma_{*}^{2}$) of both policy at each step. This result shows that as the agent deviates from the expert's trajectory towards the perpendicular distance, the confidence decreases as the data becomes more sparse.
    (c), (d) : Demonstration feasibility comparison of MHGP-BDI and MGP-BDI. (c) Demonstration trajectory with injecting a state-dependent disturbance (MHGP-BDI) and a state-independent disturbance (MGP-BDI), and (d) sequentially depicts the level of disturbance injected at each step.}
    \label{fig:sim:dist_robust}
\end{figure}

\begin{figure}
    \centering
    \includegraphics[width=\hsize]{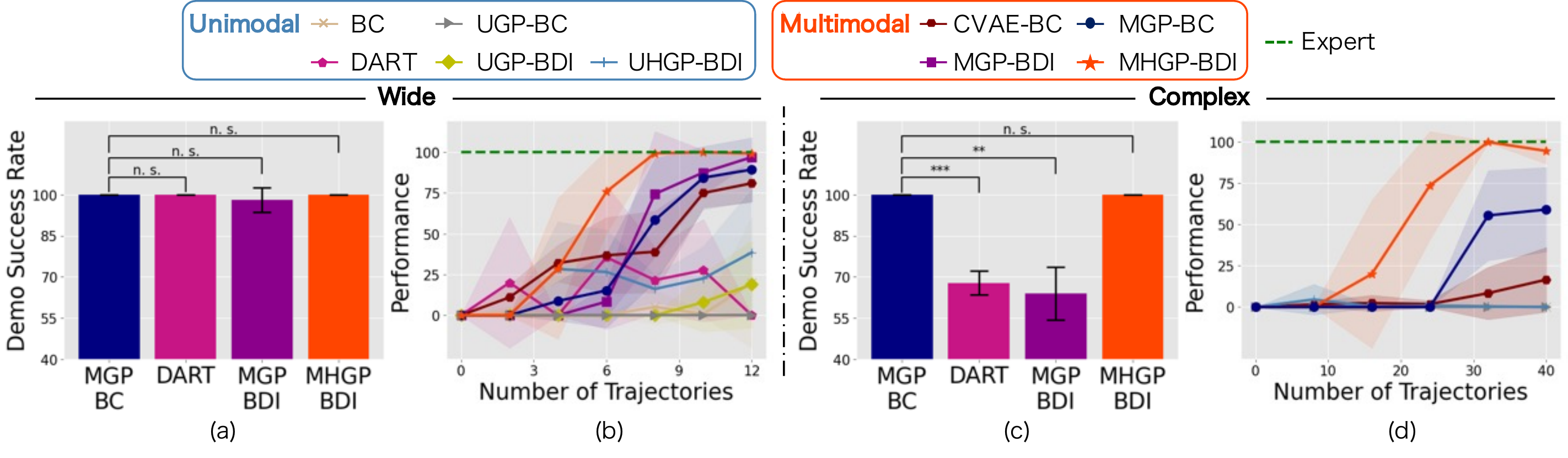}
    \vspace{-8mm}
    \caption{
    {Wall-avoidance Task Results (Wide \& Complex).}
    (a), (c): Comparing the demonstration success rate for representative learning methods (MGP-BC, DART, MGP-BDI, and MHGP-BDI) of each robustification method. The demonstration success rate of each comparison method is measured as the mean and standard deviation of the demonstration success probability for the entire trials of the final learning iteration.
    Significant differences by t-test were observed between the proposed method and baselines ($**:p < 0.005, ***:p < 0.0005$).
    Note, uni-modal methods exhibit similar results in these experiments. It is seen that demo success rate is more related to robustifying than flexibility; thus, these results focus on comparing between robustifying approaches.
    (b), (d): Comparing task performance with the number of trajectories. 
    The task performance of policy application is measured as the mean and standard deviation of the task success probability by conducting five learning trials and testing each learned policy 100 times.
    }
    \label{fig:sim:performance}
\end{figure}
In the wall-avoidance task environment (\fref{fig:sim:wide_env&multi}-(a)), the aim is for the agent (blue square with width and height $1.4~\mathrm{cm}$ and $1.5~\mathrm{cm}$ respectively) to move from the starting position (black cross) through one of the two apertures to the goal position (red circle) without colliding with the wall (grey square). The system state is the agent's position (\eg $x$, $y$-axis coordinates), and the action is the agent's velocity (\eg $x$, $y$-axis). 

Expert demonstrations are provided by an algorithmic supervisor, specifically human-like cautious behavior \citep{nagengast2011risk} is generated by a classical PID controller. This behavior is simulated by adjusting the agent's velocity during task execution: high velocity (high p-gain) in open regions far from apertures, and low velocity during aperture traversal, as shown in \fref{fig:sim:wide_env&multi}-(a).

\textbf{Wide:}
under these experimental parameters, demonstrations of passing through each aperture (aperture width is $5.0~ \mathrm{cm}$) is provided in sequence. If the agent collides with a wall or fails to reach the goal position within the time limit (400 steps), it is considered as a failure, and data is discarded, and the demonstration is restarted.
After collecting 2 demonstration trajectories, the data are used to optimize the policy and the disturbance until the optimization equation \eref{eq:MHGP:objective_function} converges. In contrast, learning methods that fail to collect demonstrations more than 5 times are considered a learning failure and are not included in the task performance comparison. This process is defined as one iteration of $k$ in \aref{algorithm:MHGP-BDI}, and is repeated $K$ times, adding the successful demonstrations to the training dataset and continuously updating the policy and disturbance until the fixed number of iterations is reached.
In this experiment, $K$ is empirically chosen to stop learning when the average of injected disturbance level is sufficiently small (\ie learned policy from each comparison is achieved at $K=6$). 
During the test execution stage, each element of the initial state is deviated by an additive uniform noise  
$\epsilon_{\mathbf{s}_{0}}\sim\mathcal{U}(- 0.05~\mathrm{cm}, 0.05~\mathrm{cm})$, and positions of the walls and goal remain constant.

\textbf{Complex:}
to evaluate the proposed method's scalability, a second experiment is also presented for a more complex task, as shown in \fref{fig:sim:dist_robust}-(a),(c). In this, apertures with a smaller width ($2.0~\mathrm{cm}$) is placed in the environment, and a secondary wall with four apertures is additionally placed below the first wall of the previous experiment. 
The clearance for moving the agent is smaller in the both layer apertures ($0.5~\mathrm{cm}$), requiring more precise control to avoid collision. 
Additionally, this secondary layer creates new traversal branches, inducing additional multiple optimal actions and requiring longer steps to accomplish the task.
Due to the increased task complexity, the time limitation is increased to 1500 step and the maximum number of demonstration trajectories for updating the policy and disturbance estimates is increased to 8 and the maximum number of iterations is $K=5$ (total 40 trajectories).
Additionally, during the test execution stage, each element of the initial state is deviated by the wider additive uniform noise $\epsilon_{\mathbf{s}_{0}}\sim\mathcal{U}(- 0.1~\mathrm{cm}, 0.1~\mathrm{cm})$.

\subsubsection{Results}
This section presents the qualitative and quantitative analysis of this simulation.
The qualitative analysis is presented in terms of (\romannumeral 1) flexibility, (\romannumeral 2) robustness, (\romannumeral 3) demonstration feasibility. 
In addition, the quantitative analysis is presented with previously described evaluation metrics.
The results of this simulation are shown in \fref{fig:sim:wide_env&multi}, \ref{fig:sim:dist_robust}, \ref{fig:sim:performance}.

\textbf{(\romannumeral 1) Flexibility:}
Initially, to evaluate the ability of the agent to flexibly learn in scenarios with multiple-optimal actions (\fref{fig:sim:wide_env&multi}-(a)), policies are learned for each of the comparison methods, and generated action distributions are shown in \fref{fig:sim:wide_env&multi}-(b). In this, it is seen that the uni-modal policy learned by UHGP-BDI fails to capture multiple optimal actions at the starting position (S) of the task. Note, all other uni-modal GP-based methods (UGP-BC, UGP-BDI) exhibit very similar Gaussian distributions. Specifically, as seen in \fref{fig:sim:wide_env&multi}-(c), uni-modal approaches learn a mean-centered policy from the demonstrations, resulting in an incorrect average direction and inability to reach any aperture. However, policies learned by MHGP-BDI can correctly capture the multi-modal distribution (\fref{fig:sim:wide_env&multi}-(b)) and learn the two optimal actions (\fref{fig:sim:wide_env&multi}-(c)).
Note, all other multi-modal GP-based methods (MGP-BC, MGP-BDI) exhibit very similar Gaussian mixture distributions.

\textbf{(\romannumeral 2) Robustness:}
To evaluate the effect of demonstrations on policy learning and application (\ie the test execution phase), initially the successful demonstrations from the MGP-BC method are used for policy learning. The results for applying policy learning is seen in (\fref{fig:sim:dist_robust}-(a)), where immediately the agent poorly performs the task by veering away from trained trajectory, and does not recover back to the optimal trajectory. This demonstrates the error compounding problem, whereby the lack of robustness in the learned model causes the agent to visit unexplored and unrecoverable states. This effect can be seen in (\fref{fig:sim:dist_robust}-(b)), whereby the action variance of MGP-BC is dramatically increased during policy learning, in a failed attempt to mitigate the problem. As such, the confidence of the policy learned by MGP-BC decreases monotonically after the 60~th time step and fails the task (time-limitation). Note that the other multi-modal neural network-based approach (CVAE-BC) exhibits a similar phenomenon.

In contrast, in the MHGP-BDI method, error compounding is minimised by injecting disturbances into demonstrations, thereby collecting recovery actions under conditions that drift from an optimal trajectory. Accordingly, when applying the policies learned in the MHGP-BDI method, even though the agent similarly immediately drifts, it can recover to an optimal trajectory and complete the task (\fref{fig:sim:dist_robust}-(a)). Even if there is a momentary decrease in confidence due to environmental variations, the policy exhibits a high confidence (\fref{fig:sim:dist_robust}-(b)). 
Note that confidence is relatively lower when passing through the first aperture (60~th time step) than the second aperture (440~time step), since the perpendicular distance from the expert's trajectory to the agent is larger, induced by the environmental variations (\eg random starting position and inertial effects).

\textbf{(\romannumeral 3) Demonstration Feasibility:}
Given this demonstration of flexibility and robustification, the disturbance injection approaches are then evaluated in terms of their ability to limit collisions. Specifically, the ability of methods which utilizes either a state-independent (MGP-BDI) or a state-dependent (MHGP-BDI) disturbance, is evaluated in aperture traversal.  In \fref{fig:sim:dist_robust}-(c), it is seen that state-independent methods, which do not regulate disturbance, collide with the walls, due to its constant level of disturbance (as seen in \fref{fig:sim:dist_robust}-(d)). As such, state-independent robustification (MGP-BDI) injects disturbances that are unsafe, and render this method unable to collect supervisor demonstrations, and the learning process cannot proceed any further. Note that the other state-independent approach (DART) exhibit similar phenomenon. In contrast, MHGP-BDI equipped with a state-dependent disturbances, successfully navigates the tasks-space, by reducing the level of disturbance to about $23\%$ of that of the MGP-BDI when it comes close to aperture ($t=86$) (\fref{fig:sim:dist_robust}-(d)). This cautious-like behavior enables the agent to pass through the aperture safely, and complete the demonstrations.

\textbf{Quantitative Evaluation:}
To evaluate the stability of these approaches, these experiments were repeated five times. The averaged demo success probabilities for representative learning methods (MGP-BC, DART, MGP-BDI, and MHGP-BDI) of each robustification method are shown in \fref{fig:sim:performance}-(a), (c). In addition, the averaged task execution performance of each learned policy is shown in \fref{fig:sim:performance}-(b), (d).

In the wide aperture experiments, (\fref{fig:sim:performance}-(a)), the demonstration feasibility is not significantly different from MGP-BC even with the disturbance injection learning approaches (DART, MGP-BDI and MHGP-BDI), since the aperture size is sufficiently large. However, (\fref{fig:sim:performance}-(b)), the uni-modal policy schemes (BC, DART, UGP-BC, UGP-BDI and UHGP-BDI) all fail to learn the multi-modal task and as expected produce low performance (under $50\%$) results, due to lack of flexibility (as discussed in \fref{fig:sim:wide_env&multi}-(b), (c)).
Note, DART, UGP-BDI, and UHGP-BDI gain additional robustness over the standard uni-modal approaches; since some deviated states, induced by control errors due to failure to capture multiple optimal actions, may be covered by disturbance injection.
Thus its performance increases monotonically in the early stages; however it eventually cannot exceed $50\%$ due to the limitation of learning flexibility. In comparison, the multi-modal policy schemes (CVAE-BC, MGP-BC, MGP-BDI and MHGP-BDI) improve the learning performance by nearly $100\%$ with increasing number of trained trajectories.

In the complex aperture experiments, (\fref{fig:sim:performance}-(d)), even multi-modal BC approaches (CVAE-BC, MGP-BC) using a flexible multi-modal policy, learned policies' task performance cannot exceed $60\%$, due to the lack of robustness (as discussed in \fref{fig:sim:dist_robust}-(a), (b)). 
However, if disturbances are injected into demonstrations in a state-independent manner (DART, MGP-BDI), this perturbation may cause physical contact at narrow apertures, and lead to demonstration failure (as discussed in \fref{fig:sim:dist_robust}-(c), (d)). This failure is seen in DART and MGP-BDI; both have a low demonstration success rate in the complex simulation (\fref{fig:sim:performance}-(c)), with demonstration success decreased by $32\%$ compared to the wide-version. Accordingly, DART and MGP-BDI are removed from the comparison of learning performance in the complex aperture experiments (\fref{fig:sim:performance}-(d)), since they failed 5 times demonstrations during learning iteration.
In contrast, MHGP-BDI, which can learn a state-dependent disturbances, has a $100\%$ demonstration success rate for both simulations, and consistently shows superior learning efficiency and obtain policies with high task performance (nearly $100\%$, only very small failures due to some specific starting position or given environmental noise).

\begin{figure}
    \centering
    \includegraphics[width=1.0\hsize]{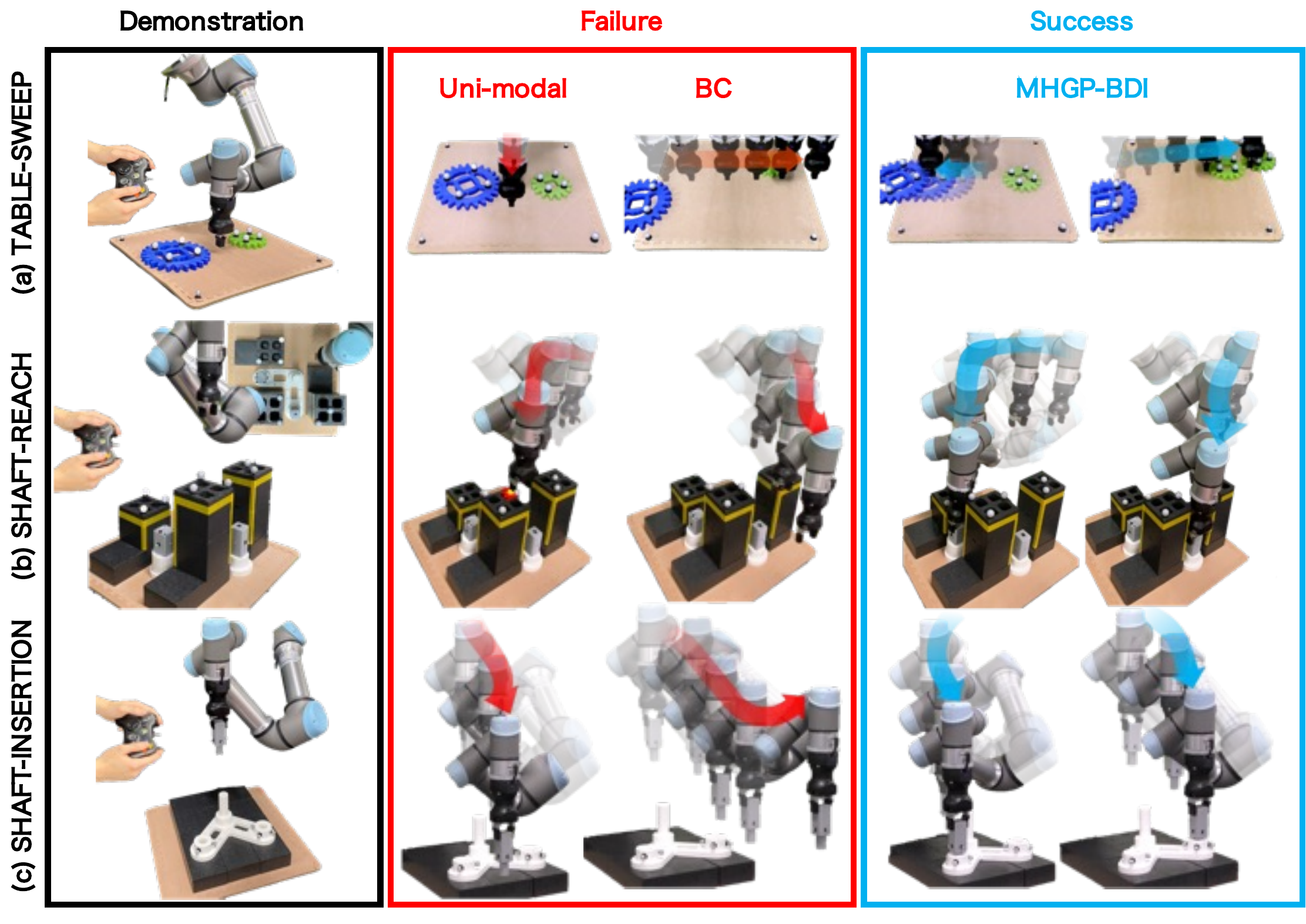}
    \vspace{-10mm}
    \caption{
    {Real Robot Experiments Setup.}
    Experimental environments for 6-DOF robotic arm (UR5e) assembly tasks with human expert are conducted as: (a) sweeping gears on the table, (b) reaching to a shaft with avoiding obstacles, (c) inserting a shaft into a hole.
    Test execution scenes of learned policies:
    $\mathbf{Failure}$: 
    (Uni-modal) 
    Due to the inability to capture the multiple optimal actions, these approaches learn mean-centred policy, resulting in (a) sweeping a centre of the gears, (b) colliding to an obstacle between shafts, (c) putting a shaft onto the centre of the holes.
    (BC) 
    Even though the approach can capture multiple optimal actions, without disturbance injection in demonstrations, policies are vulnerable to environmental variations, resulting in a robot departure from the demonstrated states; thus, the robot (a) cannot sweep gears completely or ((b), (c)) go out of the task space.
    $\mathbf{Success}$: 
    Our proposed method (MHGP-BDI) provides policies that are learned by capturing optimal actions or initiating recovery actions by injecting optimized disturbances, which allow the robot to successfully (a) sweep the whole gears, (b) reach to both shafts and (c) insert a shaft into the holes, in a given any starting position.
    Our supplementary video can be seen at: \url{https://youtu.be/NeJy8pfkrC4}.
    }
    \label{fig:real:env}
\end{figure}

\begin{figure}
    \centering
    \includegraphics[width=0.9\hsize]{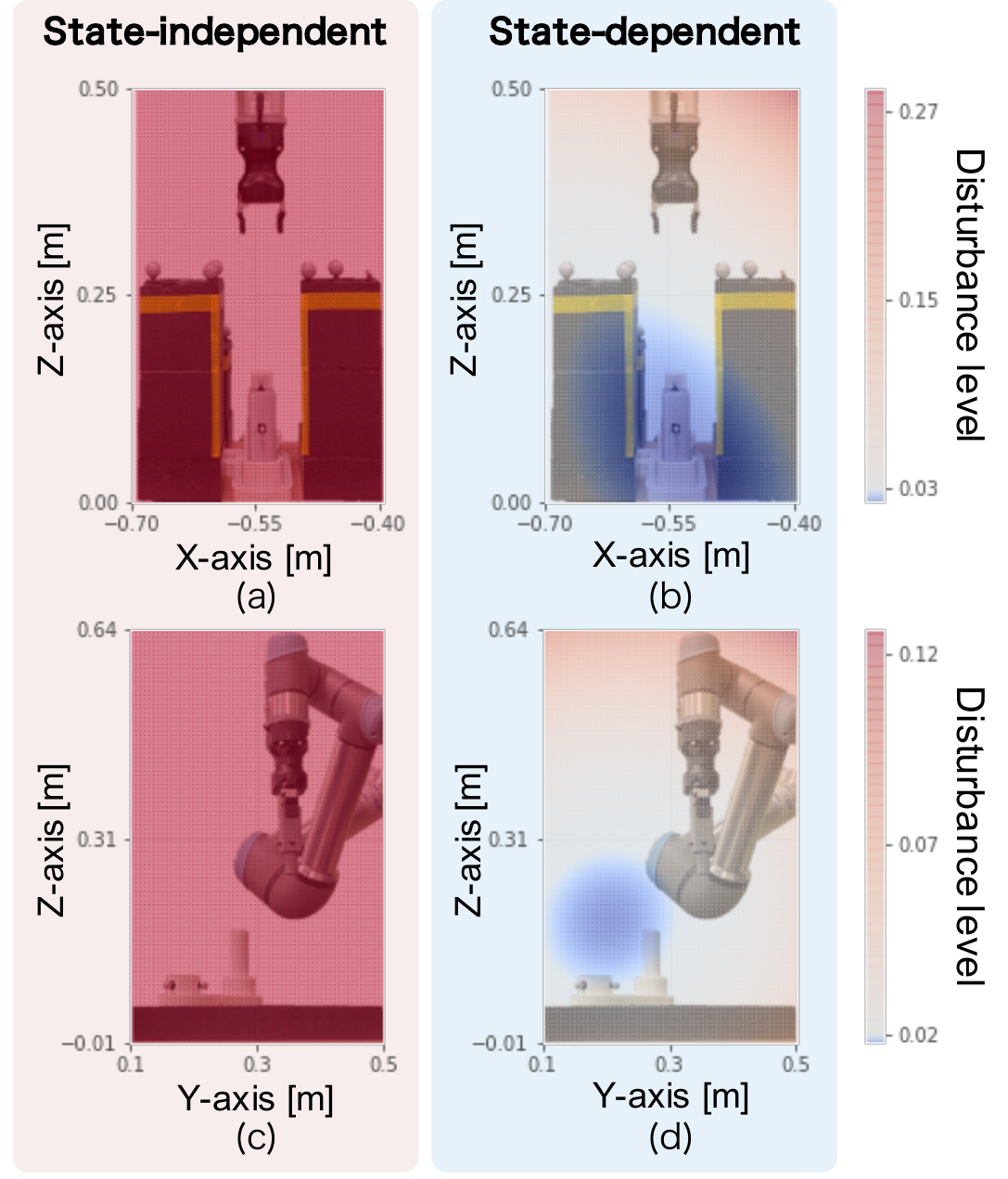}
    \vspace{-3mm}
    \caption{
    {Comparison of Two Types of Disturbances.} Both disturbances (state-independent and state-dependent) are injected into a human demonstration during a shaft-reach task ((a), (b)) and a shaft-insertion task ((c), (d)). 
    Graphs showing the disturbance level with regards to the end-effector position with fixed y-coordinate ((a), (b) fixed Y-axis = $0.23~\mathrm{m}$) and the grasped shaft position with fixed X-coordinate ((c), (d) fixed X-axis = $-0.7~\mathrm{m}$) : 
    state-independent disturbances have a uniform level in any state (left), and state-dependent disturbances, obtained by MHGP-BDI, have a spectrum of level depending on the state (right).
    Colors of disturbance level are normalized by the amount of clearances for each task.
    }
    \label{fig:real:disturbances}
\end{figure}

\begin{table*}[ht!]
\centering
\caption{
    {Real Robot Experiments Results.}
    Each learning model's demonstration success is measured at the last iteration of learning each task (10 demonstration attempts).
    If a human fails demonstration (\eg robot crashes with obstacles or fails to complete the task within the time limit) over 5 times at a single iteration then finish the learning process, failing to obtain a policy. Such learning models are not able to measure the task execution performance of learned policy; thus it is annotated as $\mathrm{N/A}$. The test execution performance of policies learned by each learning model has been measured over 10 test executions.
    }
    \label{table:real:results}
\vspace{2mm}{}
\begin{tabular}{|p{2.2cm}|C{1.35cm}|C{1.35cm}|C{1.55cm}|C{1.35cm}|C{1.35cm}|C{1.55cm}|}
\hline
\multirow{2}{*}{Learning}  & \multicolumn{3}{c|}{Demonstration Success} & \multicolumn{3}{c|}{Test Execution Performance} \\ \cline{2-7}
\multirow{2}{*}{Models}   & Table-sweep & Shaft-reach & Shaft-insertion & Table-sweep & Shaft-reach & Shaft-insertion \\ \hline\hline
BC   & 10/10 & 10/10 & 10/10 & 0/20  & 0/10  & 0/10  \\ \hline
DART   & 10/10 & 4/10 & 4/10 & 0/20  & $\mathrm{N/A}$  & $\mathrm{N/A}$  \\ \hline
CVAE-BC   & 10/10 & 10/10 & 10/10 & 16/20  & 4/10  & 5/10  \\ \hline
UGP-BC   & 10/10 & 10/10 & 10/10 & 0/20  & 0/10  & 0/10  \\ \hline
UGP-BDI  & 10/10 & 0/10 & 0/10 & 0/20  & $\mathrm{N/A}$   & $\mathrm{N/A}$   \\ \hline
UHGP-BDI & 10/10 & 10/10 & 10/10 & 0/20  & 0/10  & 0/10  \\ \hline
MGP-BC   & 10/10 & 10/10 & 10/10 & 10/20 & 7/10  & 5/10  \\ \hline
MGP-BDI  & 10/10 & 2/10 & 1/10 & 20/20 & $\mathrm{N/A}$   & $\mathrm{N/A}$   \\ \hline
\rowcolor[HTML]{C0C0C0}
MHGP-BDI (Proposed)&\vspace{-0.5mm}{10/10} & \vspace{-0.5mm}{10/10} & \vspace{-0.5mm}{10/10} & \vspace{-0.5mm}{20/20} & \vspace{-0.5mm}{10/10} & \vspace{-0.5mm}{10/10} \\ \hline
\end{tabular}
\end{table*}

\section{Real Robot Experiments}\label{sec:real_experiments}
In this section, three experiments are conducted to demonstrate the proposed method's applicability on various scenarios as shown in \fref{fig:real:env}.
MHGP-BDI is applied to a 6-DOF UR5e (Universal Robotics) robot to learn three assembly tasks:
\begin{itemize}
    \item \textbf{Table-sweep task}: 
    the robot's ability to reach multiple objects and sweep them out of the table, is evaluated. Demonstrations of sweeping two gears on the table are provided by a human, as shown in \fref{fig:real:env}-(a).
    The state of the system is defined as the relative 2D coordinate from the robotic arm to two gears ($Q=4$); an action is defined as the velocity of the robotic arm in the $x$ and $y$ axis.
    
    \item \textbf{Shaft-reach task}: 
    the robot's ability to avoid fixed obstacles and reach a shaft to grasp it, is evaluated. Demonstrations of reaching one of the assembly supplies (\eg shaft) without colliding with fixed obstacles are provided by a human, as shown in \fref{fig:real:env}-(b). 
    The state of the system is defined as the relative 3D coordinate between the robot arm and two shafts ($Q=6$), an action is defined as the velocity of the robot arm in the $x$, $y$ and $z$ axis.
    This is a more difficult task than the table-sweep task, as:
    \begin{enumerate*}[(1)]
        \item the state-action space is larger to deal with a more general setting,
        \item the environment is prone to physical contact (\eg collision with obstacles).
    \end{enumerate*}
    
    \item \textbf{Shaft-insertion task}: 
    the robot's ability for inserting a shaft into a hole for assembly, is evaluated. Demonstrations of inserting the assembly supplies (\eg shaft) into one of the holes (on both side of white ``L'' shaped base) are provided by a human, as shown in \fref{fig:real:env}-(c).
    The state of the system is defined as the relative 3D coordinate between the robot arm and two holes ($Q=6$), an action is defined as the velocity of the robot arm in the $x$, $y$ and $z$ axis. 
    This scenario is more complicated than the shaft-reach task as: 
    \begin{enumerate*}[(1)]
        \item Physical contacts are involved, requiring more sensitive behavior, 
        \item the clearance for inserting shaft is smaller (only $1~\mathrm{mm}$), requiring more precise control.
    \end{enumerate*}
\end{itemize}
In the following experiments, learned policies are evaluated in terms of ability to flexibly learn tasks with multiple optimal actions (\eg the order in which to interact with the objects), and well as robustness to environmental covariance shift inducing disturbances (\eg friction between the objects and environment, inducing variations in movement). 
Here, the test execution performance of the learned policies is measured by 10 deployment tests of the final learned policy for each learning method.
In addition, human demonstrations are evaluated in terms of feasibility for completing a task. For example, if the robot collides with obstacles or fails to complete a task during the demonstration stage within the time limit (400 steps), it is considered a failure, and the demonstration is instead repeated. 
Suppose a human fails demonstration over 5 times at a single learning iteration. In that case, it is considered learning failure (terminate the learning process), and such learning methods are removed from the task performance comparison.
Here, the demonstration success is measured by conducting 10 demonstration attempts with same conditions (\eg disturbance model) used at the last learning iteration.

To measure the state of the system, markers are attached to each object (gear, shaft, hole) and tracked through a motion capture system (OptiTrack Flex13).
In addition, to validate the robustness of the policy to deviations from optimal trajectories, each element of the initial state is deviated with additive uniform noise:
\begin{enumerate*}[(1)]
    \item table-sweep task: $\epsilon_{\mathbf{s}_{0}}\sim\mathcal{U}(- 0.05~\mathrm{m}, 0.05~\mathrm{m})$
    \item shaft-reach/insertion task: $\epsilon_{\mathbf{s}_{0}}\sim\mathcal{U}(- 0.005~\mathrm{m}, 0.005~\mathrm{m})$
\end{enumerate*}
The assembly model used \citep{siemens2017} (\eg gears, shafts, base) is a standardized benchmark task for robotic assembly. 

\subsection{Table-sweep Task}

\subsubsection{Setup}
Initially, two gears and the robot arm are placed at fixed coordinates on a table. The human expert performs demonstrations in which the objects are swept off the table. Two demonstrations from these initial conditions are then performed, capturing both variations in the order of which the objects are swept from the table. The method optimizes a policy and disturbances until \eref{eq:MHGP:objective_function} is converged. This process is repeated $K=4$ times (8 trajectories).

The learned policies' performance is evaluated according to the number of gears swept out of the table at the end of the test execution episode.

\subsubsection{Result}
The results of this experiment are seen in \tref{table:real:results}.
In the table-sweep task, the expert can successfully perform demonstrations using any of the proposed methods, even when disturbances (\ie state-dependent or state-independent) are injected; since the environment does not involve any obstacles in which disturbances may induce risks (\eg collisions or confusion in decision making).

Given these successful demonstrations, task performance is then evaluated in \tref{table:real:results}. In this, it is seen that the uni-modal policy methods (BC, DART, UGP-BC, UGP-BDI, UHGP-BDI) all fail. Specifically, in terms of flexibility, it is seen that instead of capturing multiple optimal actions at the start of sweeping, instead a mean-centered policy is learned that fails to reach either gears (\fref{fig:real:env}-(a) Uni-modal failure).
As such, they have a zero task execution performance, and demonstrate a lack of flexibility. 
In comparison, the multi-modal policy methods (CVAE-BC, MGP-BC, MGP-BDI and MHGP-BDI) correctly learn that there are multiple optimal actions (\eg move to blue or green gear), and outputs actions to sweep the two gears accordingly (\fref{fig:real:env}-(a)Success).
However, while CVAE-BC and MGP-BC incorporate flexibility, it has a low task performance ($80\%$ and $50\%$, respectively). This is due the dynamic behavior of gears varying between the test execution and training due to environmental variations (\eg friction between gears and the table), thereby introducing error compounding and resulting in the robot being unable to sweep the remaining gear after the first sweep (\fref{fig:real:env}-(a) BC failure).
In contrast, while the proposed disturbance-injected methods also experiences some uncertainty, it recovers and successfully sweep gears (\fref{fig:real:env}-(a) Success); thus MGP-BDI and MHGP-BDI show greatly improved performance (both are $100\%$).

\subsection{Shaft-reach Task}
\subsubsection{Setup}
Prior to the start of a demonstration, two shafts and robot arm are placed at fixed positions between the obstacles (black blocks) on the table. 
Following the same procedure as outlined in \sref{sec:sim:setup}, the human expert performs demonstrations in which the robot arm reach to each shaft alternatively. When the robot arm collides with an obstacle, it is considered a failure. After collecting two demonstrations, a policy and disturbances are optimized until \eref{eq:MHGP:objective_function} is converged. This process is repeated $K=4$ times (8 trajectories).

The learned policies' performance is evaluated according to the success of the test execution episode, determined by whether the robot arm grasped the shaft at the end of the episode.

\subsubsection{Result}\label{sec:exp:reach:result}
The results of this experiment are seen in \tref{table:real:results}.
In this, it is seen that the state-independent disturbance injection methods (DART, UGP-BDI and MGP-BDI) have a poor demonstration success rate, $40 \%$, $0 \%$ and $20 \%$ respectively.
Specifically, to examine this result, the learned disturbance is visualised in the state-space (\fref{fig:real:disturbances}-(a), (b)). In this, it is seen that state-independent methods generate disturbances with a uniform level, and as such inducing physical contacts (\eg collide with obstacle) at the demonstration.
In contrast, a state-dependent disturbance injection methods (UHGP-BDI and MHGP-BDI) can regulate disturbance level small when robot arm close to obstacles (\fref{fig:real:disturbances}-(b)), both have a $100 \%$ demonstrations success rate.

At the policy execution phase, it is seen that the uni-modal policy methods (BC, UGP-BC, UHGP-BDI) both fail to correctly learn policies to account for multiple optimal actions in the environment; thus robot collide with obstacle between the two shafts as shown in \fref{fig:real:env}(b)-Uni-modal failure.
As such, they have a $0 \%$ success rate, and demonstrate a lack of flexibility. 
In contrast, the multi-modal policy methods (CVAE-BC, MGP-BC and MHGP-BDI) show improved performance ($40\%$, $70\%$ and $100\%$, respectively). However, it is clear that even when incorporating flexibility, the success rate for BC is poor; since environmental variation (\eg starting position), the robot may deviate from trained states (\fref{fig:real:env}(b)-BC failure), demonstrating a lack of robustness.

\begin{table*}[h!]
\centering
\caption{
    {Shaft-insertion Task Results (Multiple Subjects).}
    Experimental results of robotic shaft insertion varied by four expert subjects. To obtain sufficient expert demonstrations, test subjects are practiced velocity control and make a smooth trajectory with simple instructions (\eg move the shaft from the starting point to the hole while sequentially decelerating the robotic arm), before performing the demonstrations. Each multi-modal approach (MGP-BC, MGP-BDI, and MHGP-BDI) has been validated during the demonstration and test execution phases. The success rate of each learning model is the mean and standard deviation of the results from four subjects.
}
    \label{table:real:multi-subjects}
\begin{tabular}{|C{1.7cm}|C{1.35cm}|C{1.5cm}|C{1.35cm}|C{1.8cm}|C{1.35cm}|C{1.6cm}|}
\hline
\multirow{3}{*}{Subjects}  & \multicolumn{3}{c|}{Demonstration Success} & \multicolumn{3}{c|}{Test Execution Performance} \\ \cline{2-7}
   & \vspace{-3mm}{MGP} & \vspace{-3mm}{MGP} & \vspace{-3mm}{MHGP} & \vspace{-3mm}{MGP} & \vspace{-3mm}{MGP} & \vspace{-3mm}{MHGP} \\
   & BC & BDI & BDI & BC & BDI & BDI \\ \hline\hline
\#1 & 10/10 & 1/10 & 10/10 & 5/10 & $\mathrm{N/A}$ & 10/10  \\ \hline
\#2 & 10/10 & 0/10 & 10/10 & 3/10 & $\mathrm{N/A}$ & 9/10   \\ \hline
\#3 & 10/10 & 1/10 & 10/10 & 6/10 & $\mathrm{N/A}$ & 10/10 \\ \hline
\#4 & 10/10 & 0/10 & 10/10 & 2/10 & $\mathrm{N/A}$ & 9/10 \\ \specialrule{.1em}{.05em}{.05em}
Success Rate ($\%$) & \vspace{-1mm}$100 \pm 0$  & \vspace{-1mm}$5.0 \pm 5.0$ & \cellcolor[HTML]{C0C0C0} \vspace{-1mm}$100 \pm 0$ & \vspace{-1mm}$40.0 \pm 15.8$ & \vspace{-1mm}$\mathrm{N/A}$ &\cellcolor[HTML]{C0C0C0} \vspace{-1mm}$95.0 \pm 5.0$ \\ \hline
\end{tabular}
\end{table*}
\subsection{Shaft-insertion Task}

\subsubsection{Setup}
Before the start of a demonstration, the ``L" shaped base and shaft grasped robot arm are placed at fixed starting positions in the environment.
This task involves physical contact (\eg between shaft and base) and requires a scheme to protect the experimental environment, including a robot and objects. 
As such, an impedance control \citep{duchaine2007impedance} is implemented, that cancels the force by adding reverse direction velocity when the shaft collides with the base.

Following the same procedure as outlined in \sref{sec:sim:setup}, the human expert performs demonstrations in which the robot arm inserts the shaft into each hole alternatively.
After collecting four demonstration, a policy and disturbances are optimized until \eref{eq:MHGP:objective_function} is converged. This process is repeated $K=3$ times (12 trajectories).

The learned policies' performance is evaluated according to the success of the test execution episode, determined by whether the shaft is in the hole at the end of the episode.

\subsubsection{Result}
The results of this experiment are seen in \tref{table:real:results}.
In the demonstration phase, methods that employ state-independent disturbance injections (DART, UGP-BDI and MGP-BDI) have a uniform strong level of disturbance in any state (seen \fref{fig:real:disturbances}-(c)). 
This disturbances make it challenging to insert the shaft; thus leading to a poor demonstration success rate ($40\%$ , $0\%$ and $10\%$, respectively).
In contrast, a state-dependent disturbance injection methods (UHGP-BDI and MHGP-BDI) regulates the disturbance level when the shaft is close to the hole (\fref{fig:real:disturbances}-(d)), and as such has a superior demonstrations success rate (both are $100\%$). This allows for both enriching the demonstrations in clear open spaces, and allowing for precision manipulation in tasks that require fine control, such as physical contact.

At the test execution of learned policies, it is seen that, as expected, the uni-modal policy methods (BC, UGP-BC, UHGP-BDI) learned mean-centered policies that generate movements between the two holes and fail the task (\fref{fig:real:env}(c)-Uni-modal failure); they have a $0 \%$ success rate, demonstrating a lack of flexibility. 
Furthermore, similar to \sref{sec:exp:reach:result},  incorporating flexibility without robustification (CVAE-BC and MGP-BC), causes the robot to deviate from trained states (\fref{fig:real:env}(c)-BC failure), giving a poor success rate ($50\%$ and $50\%$, respectively).
In contrast, policies learned by MHGP-BDI can output multiple optimal actions while robust to sources of error as shown in \fref{fig:real:env}. In particular, despite the small clearance in the hole's vicinity, it is seen that the robot can overcome with precise control, resulting in improved performance ($100\%$).

In addition, to evaluate intersubject robustness of the methods, four human experts with experience in robotics are used to compare multi-modal approaches (MGP-BC, MGP-BDI, MHGP-BDI); with results shown in \tref{table:real:multi-subjects}. 
Note, for the sake of simplicity and fairness of analysis, this experiment is conducted between GP-based multi-modal imitation learning approaches.
In this, injecting state-independent disturbances into demonstrations results in demonstration-infesibility for all subjects; thus, MGP-BDI has a poor demonstration success rate ($5.0 \pm 5.0\%$).
Test execution performance of MGP-BC similarly demonstrates poor average success rate ($40 \pm 15.8\%$), due to error compounding similar to previous experiments. However, these results show a higher intrasubject variance, due to the inherent differences between human-specific strategies.
In contrast, MHGP-BDI consistently obtains a superior success rate on both demonstrations and test executions ($100 \pm 0\%$ and $95 \pm 5.0\%$, respectively) with multiple subjects.

\section{Discussion}\label{sec:discussion}
As demonstrated in the experimental results, our proposed method of combining flexibility, robustification, and risk-sensitivity is effective for learning robust multi-action policies. 
By introducing a state-dependent disturbance, our proposed method automatically adjusts the level of disturbance to be appropriate depending on the state and can collect richer demonstration datasets, including recovery actions under challenging situations without losing demonstration feasibility.
Furthermore, several possibilities for extending the proposed method to address other significant robot learning challenges are discussed in this section.

\subsection{Exploiting Other Human Characteristics}
In regards to the overarching conceptual idea of learning \textit{human behavioral characteristics} as a fundamental part of imitation learning of robotic tasks, the proposed framework is suitable for modelling inherent behaviors, and appropriately utilizes them. BDI is a specific implementation of our proposal which models this by injecting disturbance into an expert's demonstration to learn robust multi-optimal policies within a Bayesian framework. Experimental results show that BDI significantly outperforms comparative methods, mitigating the contradictions between the assumptions of standard imitation learning algorithms and actual demonstrator behavior.

In the future, BDI can be extended and applied to mitigate not only the contradictions presented in this paper but also the following contradictions:
While the standard imitation learning assumes that the demonstrator is capable of outputting the optimal behavior in any given state, in robotic tasks that require high precision, such as a needle threading task \citep{jourdan2004needle}, even a human demonstrator rarely succeeds at one time without any mistakes. Applying imitation learning in such tasks requires a lot of time and cost for collecting demonstration data. To alleviate this contradiction, human demonstrations can be parameterized with weighted values of task achievement and enabling to learn from demonstration data that contains mistakes \citep{brown2019failure, chen2020failure,tahara2022disturbance}.

In addition, while standard imitation learning only deals with low-level control abilities, such as determining velocity from a given robot's position, many tasks in daily human life have long-term processes and are divided into symbolic sub-tasks, which require high-level planning ability to determine the sequence of sub-tasks \citep{fikes1971strips}. In applying conventional imitation learning to such tasks, even changing one sub-task requires re-learning the entire task from the beginning, which is inefficient and imposes a heavy burden on a human demonstrator. To address this contradiction, the hierarchical policy model \citep{fox2019multi, xu2019regression}, which can simultaneously execute high-level planning in symbolic task space and low-level control in geometric task space, can be employed to learn a human high-level planning ability.

\subsection{Improving Computational Efficiency}
Since the main purpose of our experiment is to investigate the effect of our method on several tasks with underlying contradictions in the demonstration data, only the Gaussian Processes (GPs) are employed as an inference tool for generating policies or disturbance of BDI.
While GP allows several advantages to our method, it still suffers from computational complexity, which significantly increases with the number of data points $N$ as \tref{table:computation}. 
As such, BDI as applied to long-term tasks with real-time control systems is limited by this underlying policy generation method.
To address this, GPs' kernels can be approximated with randomized Fourier features from the fastfood algorithm \citep{le2013icml}, which lowers the computational time of computing the inverse kernel matrix from $O(N^3)$ to $O(NW^3)$, where $W$ is a dimension of feature space.

\subsection{Dealing with Environmental Uncertainty}
While environmental uncertainty is not directly addressed in this paper, uncertainty and fuzziness of environment are another important challenges in applications to real systems. In the field of adaptive control, to cope with various uncertainties on environment, attempts to achieve more accurate control \citep{xin2022online, zhuang2022iterative} or safe-conscious engineering \citep{cheng2021asynchronous} commonly propose a probabilistic model to capture complex system dynamics' uncertainty or adapt a dynamics model to an unknown environment through iterative online learning. In light of this, BDI can be extended for employing the Model Predictive Control (MPC)-type policy model \citep{pereira2018mpc}, in which a sequence of states and actions are estimated with a stochastic dynamics model at each time step; thereby, enabling BDI to account for environmental uncertainty.

\section{Conclusion}\label{sec:conclusion}

This paper presents a novel paradigm on imitation learning, by focusing on learning human behavioral characteristics, and demonstrating its importance and usage. Our proposal Bayesian imitation learning framework injects risk-sensitive disturbances into an expert's demonstration to learn robust multi-action policies. This framework captures intrinsic human behavioral characteristics and allows for learning reduced covariate shift policies by collecting training data on an optimal set of states without losing demonstration feasibility. The effectiveness of the proposed method is verified on several simulations and real robotic tasks with human demonstrations. 

\section*{Acknowledgments}
This paper is based on results obtained from a project, JPNP20006, commissioned by the New Energy and Industrial Technology Development Organization (NEDO).

\appendix

\section{Update laws of variational posteriors $q$}{\label{appendix:update_q}}
The analytical solution of variational posterior $q^{*}(\mathbf{f}^{(m)})$ is given by the following derivation:
\begin{align}
\log q^{*}(\mathbf{f})
=\int \Big\{&\log p(\mathbf{a}^*, \mathbf{g}, \{\mathbf{f}^{(m)}\}, \mathbf{Z}, \mathbf{v} \mid \mathbf{S}; \btheta)\Big\}\cdot \nonumber\\ 
&q(\mathbf{g}) q(\mathbf{Z})q(\mathbf{v}) \dd\mathbf{g} \dd \mathbf{Z} \dd \mathbf{v}+ \mathrm{Const},\label{eq:append:log_q_f}
\end{align}
where, $\mathrm{Const}$ is a constanct term for normalizing distributions.
Accordingly, $q^{*}(\mathbf{f}^{(m)})$ is obtained by solving \eref{eq:append:log_q_f} as:
\begin{align}
q^*(\mathbf{f}^{(m)})&=\mathcal{N}(\mathbf{f}^{(m)} \mid \boldsymbol{\mu}_{\mathbf{f}}^{(m)}, \mathbf{C}^{(m)}),  \label{Appendix:F}\\
\boldsymbol{\mu}_{\mathbf{f}}^{(m)} &=\mathbf{C}^{(m)} \mathbf{B}^{(m)} \mathbf{a}^*, \\
\mathbf{C}^{(m)} &=(\Kbf^{{-1}(m)}+\mathbf{B}^{(m)})^{-1}, \\
\mathbf{B}^{(m)} &=\operatorname{diag}\{r_{nm}/\mathbf{H}_{n n}\},\\
\mathbf{H} &= \operatorname{diag}\{\exp([\boldsymbol{\mu}_{\mathbf{g}}]_{n}-[\boldsymbol{\Sigma}_{\mathbf{g}}]_{nn}/2)\}.
\end{align}

As similar to \eref{eq:append:log_q_f}, the analytical solution of variational posterior $q^{*}(\mathbf{Z})$ is given by the following derivation:
\begin{align}
\log q^*(\mathbf{Z})
=\int \Big\{&\log p({\mathbf{a}^*}, \mathbf{g}, \{\mathbf{f}^{(m)}\}, \mathbf{Z}, \mathbf{v})\Big\}\cdot \nonumber\\
&q(\mathbf{f}) q(\mathbf{g})q(\mathbf{v}) \dd\mathbf{g} \dd \mathbf{f} \dd \mathbf{v} +\mathrm{Const}.\label{eq:append:log_q_z}
\end{align}
Therefore, $q^{*}(\mathbf{Z})$ is obtained by solving \eref{eq:append:log_q_z} as:
\begin{align}
q^*(\mathbf{Z}) &= \prod_{n=1}^{N} \prod_{m=1}^{\infty} r_{nm}^{\mathbf{Z}_{nm}},\\ 
r_{nm} &=\frac{\rho_{nm}}{\sum_{m=1}^{\infty} \rho_{nm}}, \\
\log \rho_{nm}
&= - \frac{1}{2 \mathbf{H}_{n n}} \{ ({a^*_n}-[\boldsymbol{\mu}_{\mathbf{f}}^{(m)}]_n)^2 + [\mathbf{C}^{(m)}]_{nn} \}\nonumber\\
& ~~\quad{-\frac{1}{2}}\log{(2\pi \mathbf{H}_{n n})} - \psi(\alpha_{m}+\beta_{m})\nonumber\\
& ~~\quad+ \psi(\alpha_{m}) + \sum_{j=1}^{m-1}\{\psi(\beta_{j})-\psi(\alpha_{j}+\beta_{j})\},
\end{align}
where, $\psi(\cdot)$ is the digamma function.

As well as, the analytical solution of variational posterior $q^{*}(\mathbf{v})$ is given by the following derivation:
\begin{align}
\log q^*(\mathbf{v})
=\int \Big\{&\log p({\mathbf{a}^*}, \mathbf{g}, \{\mathbf{f}^{(m)}\}, \mathbf{Z}, \mathbf{v})\Big\}\cdot \nonumber\\
&q(\mathbf{f}) q(\mathbf{g})q(\mathbf{Z}) \dd\mathbf{g} \dd \mathbf{f} \dd \mathbf{Z} +\mathrm{Const}.\label{eq:append:log_q_v}
\end{align}
As such, $q^{*}(v_m)$ is obtained by solving \eref{eq:append:log_q_v} as:
\begin{align}
q^*(v_m) &= \Beta(v_{m} \mid \alpha_m, \beta_m), \label{Appendix:v}\\
\alpha_m &= 1+\sum_{n=1}^{N} r_{n m}, \\
\beta_m &= \beta + \sum_{j=m+1}^{\infty} \sum_{n=1}^{N} r_{n j},
\end{align}
where, $\Beta$ is the beta function.

\section{Lower bound of marginal likelihood $\mathcal{L}(q,\Omega^\prime)$}{\label{appendix:lowerbound}}

The lower bound of the marginal likelihood $\mathcal{L}(q,\Omega^\prime)$ is analytically obtained by the following derivation:
\begin{align}
    &\mathcal{L}(q,\Omega^\prime) \nonumber \\
    &=\sum_{m=1}^{\infty}\log \mathcal{N}(\mathbf{a}^{*} \mid \mathbf{0}, \Kbf^{(m)}+\mathbf{B}^{-1(m)})\nonumber \\
    &\quad+\sum_{n=1}^{N}\sum_{m=1}^{\infty} \Big[r_{n m} \Big\{ \psi(\alpha_m) - \psi(\alpha_m+\beta_m) - \frac{1}{2}[\bmu_{\mathbf{g}}]_{n}\nonumber \\
    &\quad + \sum_{j=1}^{m-1}\{\psi(\beta_{j})-\psi(\alpha_{j}+\beta_{j})\} - \frac{1}{2} \log2\pi - \log r_{n m} \Big\}\nonumber \\
    &\quad - \frac{1}{2} \log\{[\mathbf{B}^{(m)}]_{n n}/2\pi\}\Big] -\sum_{m=1}^{\infty}\KL{q(v_m)}{p(v_m)}\nonumber \\
    &\quad-\KL{\N(\mathbf{g} \mid {\boldsymbol{\mu}}_{\mathbf{g}}, \mathbf{\Sigma}_{\mathbf{g}})}{\N(\mathbf{g} \mid \mu_{0} \mathbf{1}, \Kbg)},
\end{align}
where,
\begin{align}
    &\KL{q(v_m)}{p(v_m)}\nonumber\\
    &=\log \{\Beta(v_m|1,\beta)/\Beta(v_m|{\alpha}_m,{\beta}_m)\} \nonumber\\
    &+({\alpha}_m-1) \psi({\alpha}_m)+({\beta}_m-\alpha ) \psi({\beta}_m)\nonumber\\
    &+(1-{\alpha}_m+\alpha -{\beta}_m ) \psi({\alpha}_m+{\beta}_m),
\end{align}
and
\begin{align}
    &\KL{\N(\mathbf{g} \mid {\boldsymbol{\mu}}_{\mathbf{g}}, \mathbf{\Sigma}_{\mathbf{g}})}{\N(\mathbf{g} \mid \mu_{0} \mathbf{1}, \Kbg)}\nonumber\\
    &=\frac{1}{2}\Big[\log \{|\mathbf{I}+\Kbg\mathbf{\Lambda}|\}-1
    +\operatorname{tr}\{(\mathbf{I} + \mathbf{\Lambda}\Kbg)^{-1}\}\nonumber\\
    &\qquad+\mathbf{1}^{\T}\left(\mathbf{\Lambda}-\frac{1}{2} \mathbf{I}\right)^{\T} \Kbg\left(\mathbf{\Lambda}-\frac{1}{2} \mathbf{I}\right) \mathbf{1} \Big].
\end{align}

\section{Computational complexity analysis of MHGP-BDI}{\label{appendix:computation}}
As described in \tref{table:computation} and \tref{table:computation:pred}, the computational complexity of MHGP-BDI is mainly related to the number of training data sets ($N$) and the upper bound of mixtures ($M$).
To analyze the impact of $N$ and $M$ on computational complexity of MHGP-BDI in practice, \textit{optimization time}, duration for one optimization loop, and \textit{prediction time}, average time for 10-step prediction, were measured in a wide version of wall avoidance simulation task; the results show that the computational complexity of MHGP-BDI is increased as $N$ and $M$ increase, as described in \tref{table:appendix:computation}. All experiments were ran on an Intel CPU Core i9-9900 K.
\begin{table}[tb]
  \centering
  \caption{Computational complexity analysis results: optimization time and prediction time in MHGP-BDI}
  \label{table:appendix:computation}
  \vspace{2mm}
  \begin{tabular}{|c|c|cc|}\hline
    $N$ & $M$ & Optimization time [sec] & Prediction time [sec]  \\ \hline\hline
    713 & 2 & 2.3 & 0.0055 \\ \hline
    711 & 5 & 5.2 & 0.013 \\ \hline
    719 & 10 & 9.9 & 0.022 \\ \hline
    1409 & 2 & 12.3 & 0.022 \\ \hline
    1407 & 5 & 27.0 & 0.053 \\ \hline
    1415 & 10 & 49.2 & 0.10 \\ \hline
  \end{tabular}
\end{table}

\section{Hyperparameters}{\label{appendix:hyper}}
Details of hyperparameters of MHGP-BDI and all comparison models are provided as below.

\subsection{MHGP-BDI}
The hyperparameters of MHGP-BDI are as follow: $M, {\omega}_{\mathbf{f}},{\omega}_{\mathbf{g}},\mu_0,\boldsymbol{\Lambda}, \beta$. 
These hyperparameters are empirically chosen with certain heuristic methodologies in this paper. 
Such hyperparameters' selection and sensitivity analysis are described as below.

The maximum number of mixtures GPs ($M$) and the concentration level parameter of SBP ($\beta$) are both related to the flexibility of the policy model. To spread out data to multiple GPs, $\beta$ is initialized as $\beta = 100$ in all experiments. In addition, $M$ is initialized based on the computational complexity of MHGP-BDI. Such as, a larger $M$ makes it better for capturing multiple optimal behaviors from human demonstrations; however, the computational complexity of the algorithm increases as $M$ grows, as shown in \tref{table:appendix:computation}. Therefore, to ensure convergence within a reasonable time frame period, $M = 5$ in all experiments.

${\omega}_{\mathbf{f}}$ and $\omega_{\mathbf{g}}$ are parameters of kernel function to regress policy's and disturbance's latent function ( $\mathbf{f}$ and $\mathbf{g}$, respectively).
In all experiments, Radial Basis Function (RBF) kernel ($k(x, y)=\exp \{-(\|x-y\|^{2})/(2 \omega^{2})\}$), which is most commonly used kernel in GP regression \citep{rasmussen2003gaussian}, is employed for all GPs.
Each parameter is initialized using the maximum and the minimum of state as: $|\max(\mathbf{S})-\min(\mathbf{S})|$.

The initial mean of disturbance level ($\mu_0$) and the positive variational parameter ($\boldsymbol{\Lambda}$) are related to action variation of human demonstrations. In all experiments, $\boldsymbol{\Lambda} = \mathrm{diag}\{\lambda_n\}_{n=1}^{N}$ is initialized as $\lambda_n = 1/2$. In addition, $\mu_0$ is initialized with variance of actions as: $\mathrm{var}(\mathbf{a}^*)\times 0.01$.

To analyze sensitivity to hyperparameters, one-at-a-time parameter sensitivity \citep{hamby1994param_sensitivity} is employed, in which the demonstration success rate and the test execution performance are measured in the wide version of wall-avoidance simulation task with varying one parameter at a time while holding the others fixed. 
Note, to simplify analysis, $\beta$ and $\boldsymbol{\Lambda}$ are initialized as $\beta = 100$ and $\lambda_{n} = 1/2$ in all experiments.
As described in \tref{table:appendix:sensitivity}, MHGP-BDI is robust to a wide range of hyperparameters.

\begin{table}[tb]
  \centering
  \caption{MHGP-BDI hyperparameters sensitivity analysis results.
  The demonstration success probability of each learning model is measured for entire learning iteration with one learning trial. The test execution performance of policy application is measured as the task success probability by conducting one learning trial and testing each final learned policy 100 times. Success rate for all demonstrations are 100 $\%$.}
  \label{table:appendix:sensitivity}
  \vspace{2mm}
  \begin{tabular}{|c|c|c|}\hline
    Parameters & Initial Value & Test Execution Performance  \\ \hline\hline
    \multirow{3}{*}{$M$}
    & $2$  & 99\% \\ \cline{2-3}
    & $5$  & 100\% \\ \cline{2-3}
    & $10$ & 100\%\\ \hline\hline
    \multirow{2}{*}{$\omega_{\mathbf{f}}$}
    & $|\max(\mathbf{S})-\min(\mathbf{S})|\times 0.1$ & 100\% \\ \cline{2-3}
    \multirow{2}{*}{$\omega_{\mathbf{g}}$}
    & $|\max(\mathbf{S})-\min(\mathbf{S})|$   & 100\% \\ \cline{2-3}
    & $|\max(\mathbf{S})-\min(\mathbf{S})|\times 10$ & 70\%\\ \hline\hline
    \multirow{3}{*}{$\mu_0$}
    & $\mathrm{var}(\mathbf{a}^*)\times 0.001$ & 100\% \\ \cline{2-3}
    & $\mathrm{var}(\mathbf{a}^*)\times 0.01$    & 100\% \\ \cline{2-3}
    & $\mathrm{var}(\mathbf{a}^*)\times 0.1$  & 100\%\\ \hline
  \end{tabular}
\end{table}

\subsection{Other GP-based comparisons}
The hyperparameter of other GP-based comparisons are described in \tref{table:appendix:gp_hyper}. 

\begin{table*}[ht!]
\centering
\caption{
    {Hyperparameters of other GP-based comparisons.}
    Since these methods employ the GP model, parameters are selected the same as in MHGP-BDI, but if the parameters are not available in the implementation, it is annotated as $\mathrm{N/A}$.
    }
    \label{table:appendix:gp_hyper}

\begin{tabular}{|p{2.2cm}|C{0.5cm}|C{0.7cm}|C{3.3cm}|C{0.7cm}|C{2.4cm}|C{0.7cm}|}
\hline
{Learning}  & \multicolumn{6}{c|}{Hyperparameters} \\ \cline{2-7}
Models & $M$ & $\beta$ & ${\omega}_{\mathbf{f}}$ & ${\omega}_{\mathbf{g}}$ & $\mu_0$ & $\lambda_n$  \\ \hline\hline
UGP-BC   & 1 & $\mathrm{N/A}$ & $|\max(\mathbf{S})-\min(\mathbf{S})|$ & $\mathrm{N/A}$ & $\mathrm{N/A}$  & $\mathrm{N/A}$  \\ \hline
UGP-BDI  & 1 & $\mathrm{N/A}$ & $|\max(\mathbf{S})-\min(\mathbf{S})|$ & $\mathrm{N/A}$ & $\mathrm{N/A}$   & $\mathrm{N/A}$   \\ \hline
UHGP-BDI & 1 & $\mathrm{N/A}$ & \multicolumn{2}{c|}{$|\max(\mathbf{S})-\min(\mathbf{S})|$} & $\mathrm{var}(\mathbf{a}^*)\times0.01$  & $1/2$  \\ \hline
MGP-BC   & 5 & 100 & $|\max(\mathbf{S})-\min(\mathbf{S})|$ & $\mathrm{N/A}$ & $\mathrm{N/A}$  & $\mathrm{N/A}$  \\ \hline
MGP-BDI  & 5 & 100 & $|\max(\mathbf{S})-\min(\mathbf{S})|$ & $\mathrm{N/A}$ & $\mathrm{N/A}$   & $\mathrm{N/A}$   \\ \hline
\end{tabular}
\end{table*}

\subsection{Neural networks-based comparisons}
The hyperparameter of neural networks-based comparisons are described in \tref{table:appendix:nn_bc_dart_hyper} and \tref{table:appendix:cvae_hyper}. 
These hyperparameters are set based on original papers \citep{laskey2017dart, ren2020generalization}, but some parameters are tuned to improve performance in our domain.

\begin{table*}[h!]
\centering
\caption{Hyperparameters of BC and DART.}
\label{table:appendix:nn_bc_dart_hyper}

\begin{tabular}{|p{6cm}|C{2cm}|}
\hline
Hyperparameter  & Value \\ \hline\hline
optimizer & Adam  \\ 
learning rate & $1\times 10^{-2}$ \\
weight decay & $1\times 10^{-5}$ \\
number of hidden layers & 2 \\
number of hidden units per layer & 64 \\
number of sample per minibatch & 128 \\
activation function & Tanh \\ \hline
\end{tabular}
\end{table*}

\begin{table*}[h!]
\centering
\caption{
    {Hyperparameters of CVAE-BC.}
    }
    \label{table:appendix:cvae_hyper}
\begin{tabular}{|p{6cm}|C{2cm}|}
\hline
Hyperparameter  & Value \\ \hline\hline
optimizer & Adam  \\ 
learning rate & $1\times 10^{-3}$ \\
weight decay & $1\times 10^{-5}$ \\
number of hidden layers & 2 \\
number of hidden units per layer & 64 \\
number of sample per minibatch & 64 \\
activation function & ReLU \\ 
number of latent dimension & 5 \\ \hline
\end{tabular}
\end{table*}

\newpage

\bibliographystyle{elsarticle-harv} 
\bibliography{reference}

\end{document}

%% file: preamable.tex
\newcommand*{\sref}[1]{Section~\ref{#1}}            
\newcommand*{\aref}[1]{Algorithm~\ref{#1}}  
\newcommand*{\tref}[1]{\tablename~\ref{#1}}   
\newcommand*{\fref}[1]{\figurename~\ref{#1}}  
\newcommand*{\eref}[1]{(\ref{#1})}            

\usepackage[table,xcdraw]{xcolor}
\usepackage{ctable}

\usepackage[inline,shortlabels]{enumitem}\setlist[enumerate]*{label=(\roman*)}

\usepackage{array}
\newcolumntype{C}[1]{>{\centering\arraybackslash}p{#1}}
\usepackage[colorlinks,bookmarksopen,bookmarksnumbered,citecolor=red,urlcolor=red]{hyperref}

\usepackage{amsmath}
\usepackage{amssymb}
\usepackage{accents} 
\usepackage{algorithm,algorithmic}
\usepackage{pifont}
\usepackage{multirow}

\newcommand{\cmark}{\ding{51}}%
\newcommand{\xmark}{\ding{55}}%

\newcommand{\eg}{\textit{e.g.,}~} %
\newcommand{\ie}{\textit{i.e.,}~} %

\DeclareMathOperator*{\argmax}{arg\,max}

\newcommand{\relmiddle}[1]{\mathrel{}\middle#1\mathrel{}}

\newcommand{\N}       {\mathcal{N}}            
\newcommand{\Beta}    {\operatorname{Beta}}            

\newcommand{\btheta}  {\boldsymbol{\theta}}    

\newcommand{\bmu}      {\boldsymbol{\mu}}       

\newcommand{\dd}      {\mathrm{d}}              
\newcommand{\T}       {\top}                    
\newcommand{\KL}[2]{\operatorname{KL}({#1}\mid\mid{#2})}

\newcommand{\traj}      {\boldsymbol{\tau}}       

\newcommand{\Kbf}      {\mathbf{K}_{\mathbf{f}}}
\newcommand{\Kbg}      {\mathbf{K}_{\mathbf{g}}}